\documentclass[final]{elsarticle}
\usepackage{enumitem}
\usepackage{lineno}
\modulolinenumbers[1]

\usepackage[utf8]{inputenc}
\usepackage[english]{babel}

\journal{Swarm and Evolutionary Computation}

\usepackage{tikz,bm,color}
\usetikzlibrary{fit}

\usepackage{amsmath}
\usepackage{amsthm}
\usepackage{amssymb}
\usepackage{mathtools}
\usepackage{siunitx}

\usepackage[linesnumbered, ruled, vlined]{algorithm2e}
\usepackage {listings}

\usepackage{caption}
\usepackage{subcaption}
\usepackage {float}

\usepackage{url}

\usepackage[all]{myglossary_minus}

\usepackage{tabularx}
\usepackage{multirow}
\usepackage{array}
\usepackage{soul}

\newcolumntype{L}[1]{>{\raggedright\let\newline\\\arraybackslash\hspace{0pt}}m{#1}}
\newcolumntype{C}[1]{>{\centering\let\newline\\\arraybackslash\hspace{0pt}}m{#1}}
\newcolumntype{R}[1]{>{\raggedleft\let\newline\\\arraybackslash\hspace{0pt}}m{#1}}

\theoremstyle{definition} 
\newtheorem{definition}{Definition}
\makeglossaries
\begin{document}

\begin{frontmatter}

\title{\mbox{Weighted Strategies to Guide a Multi-Objective} \mbox{Evolutionary Algorithm for Multi-UAV Mission Planning}}


\author[uam]{Cristian Ramirez Atencia}
\ead{cristian.ramirez@inv.uam.es}

\author[tec,ehu,bcam]{Javier Del Ser}
\ead{javier.delser@tecnalia.com}

\author[uam]{David Camacho}
\ead{david.camacho@uam.es}

\address[uam]{Departamento de Ingenier\'ia Inform\'atica, Universidad Aut\'onoma de Madrid,\\ C/Francisco Tom\'as y Valiente 11, 28049 Madrid, Spain}
\address[tec]{OPTIMA Area, TECNALIA, E-48160 Zamudio, Spain}
\address[ehu]{Department of Communications Engineering, University of the Basque Country (UPV/EHU), E-48013 Bilbao, Spain}
\address[bcam]{Basque Center for Applied Mathematics (BCAM), E-48009 Bilbao, Spain}




\begin{abstract}
Management and mission planning over a swarm of \gls*{uav2} remains to date as a challenging research trend in what regards to this particular type of aircrafts. These vehicles are controlled by a number of \gls*{gcs}, from which they are commanded to cooperatively perform different tasks in specific geographic areas of interest. Mathematically the problem of coordinating and assigning tasks to a swarm of \gls*{uav2} can be modeled as a constraint satisfaction problem, whose complexity and multiple conflicting criteria has hitherto motivated the adoption of multi-objective solvers such as \gls*{moea}. The encoding approach consists of different alleles representing the decision variables, whereas the fitness function checks that all constraints are fulfilled, minimizing the optimization criteria of the problem. In problems of high complexity involving several tasks, \gls*{uav2} and \gls*{gcs}, where the space of search is huge compared to the space of valid solutions, the convergence rate of the algorithm increases significantly. To overcome this issue, this work proposes a weighted random generator for the creation and mutation of new individuals. The main objective of this work is to reduce the convergence rate of the \gls*{moea} solver for multi-UAV mission planning using weighted random strategies that focus the search on potentially better regions of the solution space. Extensive experimental results over a diverse range of scenarios evince the benefits of the proposed approach, which notably improves this convergence rate with respect to a na\"ive \gls*{moea} approach.
\end{abstract}


\end{frontmatter}


\glsresetall

\section{Introduction}

\Glspl*{uas}, also referred to as drones, \gls*{uav2}, and \gls*{rpas}, can be defined as an aircraft without a human pilot on-board. Instead, the UAS is controlled from an operator on the ground. The main difference between \gls*{uav2} and \gls*{rpas} resides on the fact that \gls*{uav2} are basically planes that do not have a pilot, and no human intervention is made during the flight plan. \Gls*{uav2} rely on a pre-programmed flight plan, and it is the on-board computer of the aircraft what actually controls it and reacts to changing conditions in order to reach the desired location. However, in case of \gls*{rpas}, the need for a pilot is not really overridden but the pilot is just relocated to a remote location. The \gls*{rpas} still operates as a typical airplane: the pilot controls it from a cockpit with all the necessary controls, and the inputs provided by the pilot are transmitted to the remote \gls*{rpa}. This paper will focus on \gls*{uav2}, so the aircraft control will rely on a pre-programmed flight plan.

The advent of pilot-less aerial vehicles and rapid development of their capabilities have paved the way towards new military and commercial applications. \Gls*{uav2} have become very popular in many  applications including surveillance~\cite{Perez-Carabaza2016}, training~\cite{Rodriguez-Fernandez2017}, disaster and crisis management~\cite{Wu2006} and firefighting \cite{Bilbao2015}, since they avoid risking human lives while their manageability permits to reach areas of hard access. Possibilities for \gls*{uav2} span further towards the last-mile delivery of goods \cite{murray2015flying} and support for wireless communications \cite{mozaffari2015drone,zeng2016wireless}, among many others (see \cite{valavanis2014handbook} for a comprehensive survey).

The process of mission planning for a collaborative swarm of \gls*{uav2} involves generating tactical goals, commanding vehicles, risk avoidance, coordination and timing. Currently, \gls*{uav2} are controlled remotely by human operators from \gls*{gcs}, using rudimentary planning systems, such as pre-configured plans, classical planners that are not able to cope with the entire complexity of the problem, or manually provided schedules. Classical planners are based on a graph search or resort to a logic engine. Remarkably, this kind of planners usually undergo severe practical limitations, which add to the high computational resources demanded by their underlying solvers \cite{ponda2015cooperative}. When undertaking complex coordinated missions, planning systems demand more efficient problem-solving capabilities to cope with conflicting objectives and stringent constraints over both spatial and time domains. Therefore, the so-called \gls*{mcmpp} have lately gained momentum in the research community, propelled by a notable increase in the scales and number of applications foreseen for \gls*{uav2} in the near future.

Nowadays most \gls*{gcs} consider a relation between N operators and 1 \gls*{uav2} due to the high complexity of a mission. The purpose of this work is to present a novel approach to reduce this complexity, so this relation can be swapped, first to one between N operators and \gls*{uav2}, and in a near future to a single operator controlling multiple vehicles.

In the literature several algorithmic approaches have been explored and reported for \gls*{uav2} mission planning. Fabiani et al. \cite{Fabiani2007} modelled the problem for search and rescue scenarios using \gls*{mdp} and solve it with dynamic programming algorithms. Leary et al. \cite{Leary2011} compared the performance of \gls*{milp}, \gls*{sa} and \gls*{aa} for mission planning problems with multiple \gls*{uav2}. Wang et al. \cite{Wang2014} represented the task scheduling of \gls*{uav2} as a \gls*{milp} and used \gls*{ts} to solve this problem. Shang et al. \cite{Shang2014} proposed a combination of \gls*{ga} and \gls*{aco} for solving the \gls*{uav2} team oriented problems. 

An essential concept in mission planning is cooperation or collaboration, which occurs at a higher level when various \gls*{uav2} work together in a common mission sharing data and controlling actions together. There are few contributions that deal with Multi-UAV problems in a deliberative paradigm (cooperative task assignment and mission planning). Bethke et al.~\cite{Bethke2008} proposed an algorithm for cooperative task assignment that extends the \gls*{rhta} algorithm to select the optimal sequence of tasks for each \gls*{uas}. Another approach by Kvarnstrom et al. \cite{Kvarnstrom2010} proposes a new mission planning algorithm for collaborative \gls*{uav2} based on combining ideas from forward-chaining planning with partial-order planning. This approach led to a new hybrid \gls*{pofc} framework that meets the requirements on centralization, abstraction, and distribution found in realistic emergency services settings. Other works have been focused on distributed approaches for solving collaborative mission planning. Pascarell et al. \cite{PascarellaEtAl2015} proposed a \gls*{cpg} algorithm for trajectory planning. Finally, other approaches formulate the mission planning problem as a \gls*{csp} \cite{Bartak2011}, where the tactic mission is modelled and solved using \gls*{csp} techniques \cite{Guerriero2014}, in combination with genetic algorithms \cite{Ramirez-Atencia2016} or represented inside a distributed multi-agent system \cite{Hao2017,Miloradovic2017}.

Most of the above contributions assume that \gls*{uav2} are commanded by a single GCS, but in practice multi-UAV missions require the use of several \gls*{gcs} for controlling all the \gls*{uav2} involved in the mission scenario. This multi-GCS approach makes the underlying problem even more complex. Besides, several criteria can be adopted to quantify the quality of a solution, such as the fuel consumption, the makespan, the cost of the mission, or the number of \gls*{uav2} or \gls*{gcs} to employ and different risk factors that could compromise the mission, among others. In these cases, the problem is considered as a \gls*{mop}, for which an estimation of the \gls*{pof} must be inferred so as to get a portfolio of solutions (mission plans) differently albeit optimally balancing the considered conflicting objectives. To this end, an algorithmic option is to use \gls*{moea}, which are inspired from evolutionary reproduction and mutation mechanisms to efficiently construct an estimation of the \gls*{pof} for a given \gls*{mop}. The literature is scarce in what refers to this family of multi-criteria solvers when applied to multi-GCS \gls*{mcmpp}, with references dealing with path planning rather than mission scheduling (see e.g. \cite{besada2013performance} and references therein).

The vehicles considered in this work are based on Airbus Defence \& Space \gls*{uav2} \footnote{http://www.airbus.com/defence/uav.html}, as part of the SAVIER project. Four types of \gls*{uav2} have been considered: \gls*{male}, \gls*{hale}, \gls*{ucav} and \gls*{urav}, this last one is based on the EADS ATLANTE model \footnote{http://www.airforce-technology.com/projects/atlante-uav/}. All utilized vehicles have 6 degrees of freedom (DoF). In addition, the definition of the missions, the flight profiles and the following procedures have been specified by Airbus engineers involved in this project according to realistic standards in aerial simulations.

This manuscript extends and improves a previous approach dealing with a \gls*{moea} to tackle this family of problems \cite{Ramirez-Atencia2016}. In this new approach, the encoding consists of different alleles representing the features of the problem, whereas the fitness function builds upon several conflicting objectives of the application scenario, which are to be minimized subject to a number of constraints. Unfortunately, in some problems of high complexity involving several tasks, \gls*{uav2} and \gls*{gcs}, where the space of search is huge compared to the space of valid solutions, this approach is unable to converge in a reasonable time. Therefore, in order to improve it and guide the algorithm to valid solutions, this work takes a step further by proposing a novel \textit{weighted random generator} for the formation of new individuals. This generator is applied in three parts of the encoding:

\begin{itemize}[noitemsep,leftmargin=*]
\item First, we use a weighted random function that assigns lower probability to individuals with higher numbers of \gls*{uav2} for performing the tasks.
\item Second, the selection of the \gls*{uav2}(s) performing the task at hand is done by using a weighted random function driven by the distance of the \gls*{uav2}(s) to the task.
\item Third, \gls*{gcs} controlling a \gls*{uav2} is selected by means of a weighted random function depending on the distance of the \gls*{gcs} to the \gls*{uav2}.
\end{itemize}
 
To test this approach, several weighted strategies are proposed and assessed over a number of realistic scenarios with varying scales. The obtained results are promising and encourage the adoption of the proposed simple heuristics in \gls*{moea} applied to other multi-criteria mission planning scenarios.\\



The rest of the paper is structured as follows. Section \ref{formulation} provides a baseline mathematical formulation of the considered \gls*{uav2} mission planning problem, including an overview of the \gls*{csp} modelling of the problem. Section \ref{pmogammp} presents the designed solver to efficiently tackle the aforementioned problem, providing details on its encoding, constraint handling strategy and the weighted strategies used for producing and evolving individuals. Section \ref{experimentalresults} discusses the results obtained from several computer experiments using realistic data. Finally, the last section draws conclusions and outlines future research lines related to this work.

\section{System Model and Problem Formulation}\label{formulation}

The \gls*{mcmpp} can be defined as a $\texttt{T}$-sized set of \textit{tasks} $\mathcal{T}\doteq\{T_1,T_2,\ldots,T_\texttt{T}\}$ to be performed by a swarm of $\texttt{U}$ \gls*{uav2} $\mathcal{U}=\{U_1,U_2,\ldots,U_\texttt{U}\}$ within a specific time interval. Each mission should be performed in a specific geographic zone. In addition, $\texttt{G}$ \gls*{gcs} $\mathcal{G}\doteq\{G_1,G_2,\ldots,G_\texttt{G}\}$ control the swarm of \gls*{uav2}. A basic mission planning comprises the assignment of each task $T_t$ (with $t\in\{1,\ldots,\texttt{T}\}$) to a specific \gls*{uav2} $U_u$ (corr. $u\in\{1,\ldots,\texttt{U}\}$), as well as each \gls*{uav2} to a specific \gls*{gcs} $G_g$ ($g\in\{1,\ldots,\texttt{G}\}$), ensuring that the mission can be successfully performed within a time frame. Such a solution can be mathematically formulated as two many-to-one mapping functions $\lambda_{TU}: \{1,\ldots,\texttt{T}\} \mapsto \{1,\ldots,\texttt{U}\}$ and $\lambda_{UG}: \{1,\ldots,\texttt{U}\} \mapsto \{1,\ldots,\texttt{G}\}$ such that $\lambda_{TU}(t)=u$ and $\lambda_{UG}(u)=g$ denotes that task $T_t$ is assigned to \gls*{uav2} $U_u$, which in turn is commanded by \gls*{gcs} $G_g$.

The set $\mathcal{T}$ can collect very diverse tasks, from photographing to escorting a target, monitoring a zone or providing support for on-site communications. Some of such tasks can be performed by several \gls*{uav2} (Multi-UAV), hence reducing the time needed to accomplish them, e.g. mapping an area or the so-called step \& stare. Each task must be performed over a particular geographic area and within a specific time interval. Furthermore, tasks require the use of sensors installed on every UAV (i.e. \gls*{eoir} cameras, \gls*{sar}, \gls*{isar} \gls*{mpr}, etc.) participating in the mission. To perform a specific task, the \gls*{uav2}(s) assigned to perform it must use one specific sensor while on operation. Although in real scenarios, some complex task may precise the use of several sensors of the same vehicle at the same time, this special case has been omitted in this problem. This poses further constraints to the mission planning, as it might be the case that not all \gls*{uav2} in the swarm have all type of sensors installed on board, hence not being eligible to undertake any task within $\mathcal{T}$. This being said, \gls*{uav2} performing the mission have some features that must be taken into account in order to check if a mission plan is correct:
\begin{itemize}[noitemsep,leftmargin=*]
\item The initial position,in terms of latitude (LAT) and longitude (LONG), of the UAV $U_u$ (with $u\in\{1,\ldots,\texttt{U}\}$), denoted as $\mbox{Pos}_u(\tau)\doteq (\mbox{LAT}_u(\tau),\mbox{LON}_u(\tau))$, with $\tau$ denoting time.
\item The fuel of the UAV during the mission plan, denoted as $\mbox{FUEL}_u(t)$ (in kilograms).
\item The cost per hour of usage of a UAV $C_u$ (in monetary units per hour).
\item The maximum flight time and range of UAV $U_u$, measured as the $T_u^{max}$ and $D_u^{max}$.
\item The available sensors at every UAV, by which a binary correspondence $\xi: \{1,\ldots,\texttt{U}\}\times \{1,\ldots,\texttt{T}\}\mapsto \{0,1\}$ can be built such that $\xi(u,t)=1$ if UAV $U_u\in\mathcal{U}$ has all sensors required to perform task $T_t\in\mathcal{T}$ installed on board, and $\xi(u,t)=0$ otherwise. Clearly, $\lambda_{TU}(t,u)=0$ whenever $\xi(u,t)=0$ (i.e. no UAV should undertake any task for which it is unqualified). When there is more than a sensor on-board capable of performing a given mission, the selection of which sensor to use should be included as another design criterion for the plan due to the direct involvement of this variable in the cost and the maximum flight time of the UAV. Sensors can be non-cooperative (such as the cameras used in a target photographing task) or cooperative (correspondingly, radars working together in a surveillance task). These cooperative sensors are used in Multi-UAV tasks, where different vehicles cooperate and their sensors coordinate among themselves to achieve the best performance when undertaking the specific task.
\item $\texttt{F}\geq 1$ flight profiles $\mathcal{F}\doteq \{F_1,\ldots,F_\texttt{F}\}$, each specifying speed $S_f(\tau)$, fuel consumption rate in kilograms per hour $R_f^{\scriptsize{\mbox{FUEL}}}(\tau)$ and flight altitude $A_f(\tau)$ in feet (or equivalently, ascent/descent angle) during the different time instants $\tau$ of the profile. Since these operational parameters have direct implications on most of the above features (e.g. fuel consumption of the UAV, its maximum flight time and the range or maximum flight distance), flight profiles are also optimization variables to be assigned so as to complete the mission plan successfully. In this problem, four flight profiles have been considered for each \gls*{uav2}: a climb profile, used for every altitude rise; a descent profile, used for every descent of altitude; a minimum consumption profile, where the speed is low and the altitude is high, so the fuel consumption is at its minimum; and a maximum speed profile, where the speed is at its maximum value, the altitude low, and so the fuel consumption is high. 
\end{itemize}


Additionally, there could exist vehicle and time dependencies between different tasks in a mission. Vehicle dependencies reflect whether two tasks can be assigned to the same UAV or instead must be mapped to different \gls*{uav2}, i.e. whether the task-to-UAV mapping function $\lambda_{TU}(t)$ as defined above is not injective. Time dependencies pose restrictions on the start and end times of related tasks so they are performed simultaneously, consecutively or in any other mutual relation of dependency along time. Such temporal relations can be expressed in terms of the Allen's interval algebra \cite{Allen1983} as follows:
\begin{itemize}[noitemsep,leftmargin=*]
\item $T_t < T_{t'}$: $T_t$ takes place before $T_{t'}$.
\item $T_t\; m\; T_{t'}$: $T_t$ \emph{meets} $T_{t'}$ (i.e. $T_{t'}$ starts right after $T_t$).
\item $T_t\; o\; T_{t'}$: $T_t$ overlaps $T_{t'}$.
\item $T_t\; s\; T_{t'}$: $T_t$ starts jointly with $T_{t'}$.
\item $T_t\; d\; T_{t'}$: $T_t$ starts and ends while $T_{t'}$ is performed.
\item $T_t\; f\; T_{t'}$: $T_t$ finishes together with $T_{t'}$.
\item $T_t = T_{t'}$: $T_t$ is equal to $T_{t'}$.
\end{itemize}

Similarly, the set of \gls*{gcs} controlling the deployed \gls*{uav2} also feature several aspects to be considered when checking a mission plan:
\begin{itemize}[noitemsep,leftmargin=*]
\item The geographical position where each \gls*{gcs} is located, defined as $\mbox{Pos}_g\doteq (\mbox{LAT}_g,\mbox{LON}_g)$ for $g\in\{1,\ldots,\texttt{G}\}$.
\item The maximum number of \gls*{uav2} that every \gls*{gcs} $ G_g$ can control, given by $1\leq \texttt{U}_g^{max}\leq \texttt{U}$.
\item The permitted types of \gls*{uav2} that every \gls*{gcs} can command.
\item The coverage or within range of the communications equipment of every \gls*{gcs}, which is assumed to be circularly shaped with radius $R_g^{max}$ (in nautical miles).
\end{itemize}

\begin{figure}[!h]
\centering
\includegraphics[width=\columnwidth]{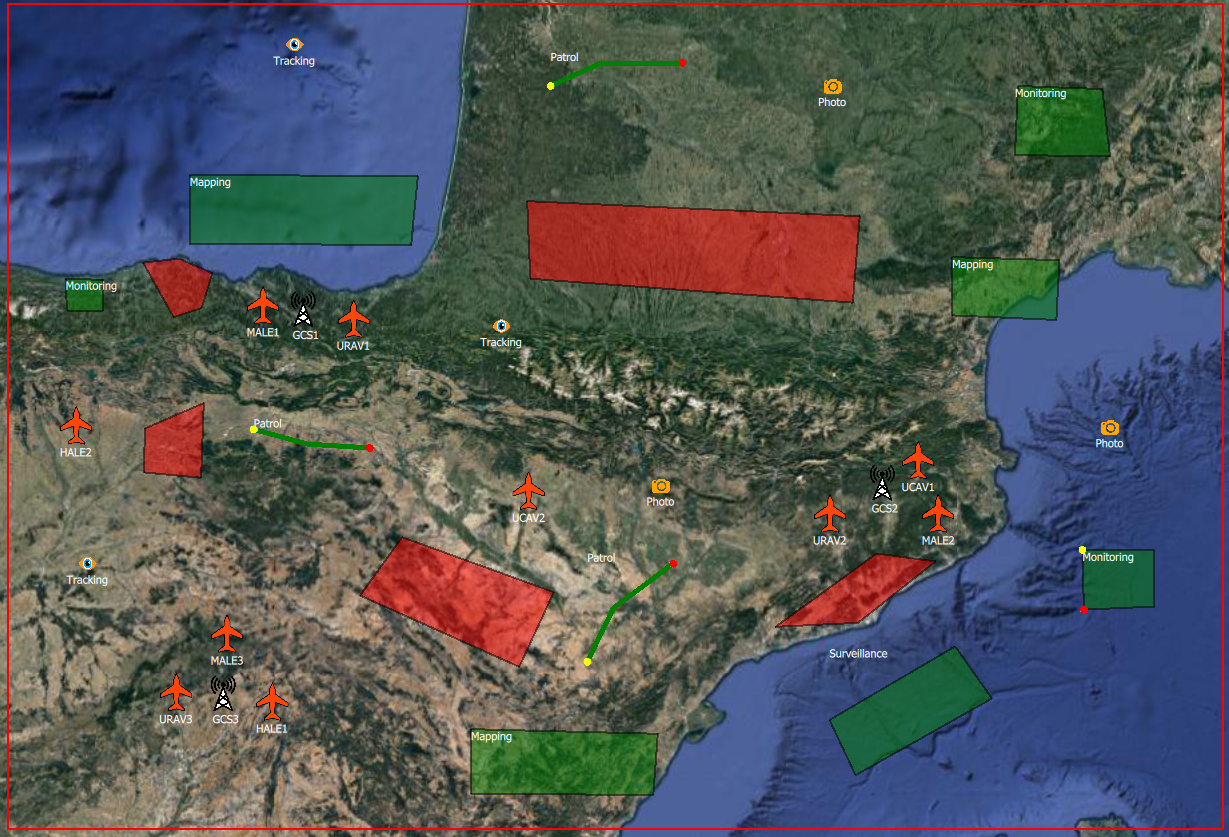}
\caption{Snapshot exemplifying a mission with $\texttt{T}=16$ tasks ($4$ of them comprising multiple UAV), $\texttt{U}=8$ UAV and $\texttt{G}=3$ GCS.}
\label{fig:bigMission}
\end{figure}

Figure \ref{fig:bigMission} presents a mission scenario with $\texttt{T}=16$ tasks  (represented as green squares and lines, and symbols as photo or tracking that denote a specific task to be done in that area), $\texttt{U}=8$ \gls*{uav2} and $\texttt{G}=3$ \gls*{gcs}. As shown in this figure, the zone of the mission contains several \gls*{nfz} shaded in red. These zones must be avoided in the planned trajectories of the \gls*{uav2} during the mission, impacting on the duration of routes traced over the scenario and ultimately, on the fuel consumption, maximum flight time, range and other parameters related to the dynamics of the application scenario.

When a task is assigned to a vehicle, it is necessary to compute the total time taken for the UAV to complete the path between its departure location and the area where the task is to be performed. If a task $T_t$ is the last one assigned to a vehicle $U_u$, the time taken to return from this last task to the base must be also calculated. In order to compute these times, it is necessary to know which of the set of \gls*{uav2}'s flight profiles $\mathcal{F}$ will be used, providing the fuel consumption ratio, speed and altitude as previously mentioned. 

Furthermore, the assignment of tasks to UAV is also subject to the order in which such tasks are accomplished during the mission plan, always in compliance with the time constraints between tasks. 
In mathematical terms we will define the mapping function $\Gamma_{TU}: \{1,\ldots,\texttt{T}\} \times \{1,\ldots,\texttt{U}\} \mapsto \{0,\ldots,\texttt{T}-1\}$, which establishes an order for every UAV $U_u\in\mathcal{U}$ to accomplish any given task $T_t\in\mathcal{T}$. It should be clear that $\Gamma_{TU}(t,u)$ exists iff (\emph{if and only if}) $\lambda_{TU}(t,u)=1$, i.e. UAV $U_u$ is assigned to perform task $T_t$. For UAV $U_u\in\mathcal{U}$ the subset of its assigned tasks will be given by $\Gamma_{U_u} = \{\Gamma_{TU}(t,u):\: \lambda_{TU}(t,u)=1\}$, such that $\cup_{u=1}^V \Gamma_{U_u} = \mbox{Perm}(|\mathcal{T}|)$, where $\mbox{Perm}(a)$ denotes a permutation of the integer set $\{0\ldots,a-1\}$. In addition, these orders must satisfy the time constraints between any pair of tasks $T_t$ and $T_{t'}$ (with $t\neq t'$). For example, if UAV $U_1$ performs tasks $T_1 \to T_5$, we could define $\Gamma_{U_1}=\{4,5,1,3,2\}$ such that $T_5<T_1$ and $T_3<T_2$.

Given the above definitions, a mission plan $\bm{\Phi}$ can be defined as 
\begin{equation}
\bm{\Phi}\doteq [\lambda_{TU},\Gamma_{TU},\lambda_{UG},\lambda_{TUF},\lambda_{TUS}],
\end{equation}
where $\lambda_{TUF}: \{1,\ldots,\texttt{T}+1\} \times \{1,\ldots,\texttt{U}\} \mapsto \{1,\ldots,\texttt{F}\}$ is a mapping function that establishes flight profiles to any UAV/task combination (including task $T_{\texttt{T}+1}=\texttt{ReturnToBase}$); and $\lambda_{TUS}: \{1,\ldots,\texttt{T}\} \times \{1,\ldots,\texttt{U}\} \mapsto \{1,\ldots,\texttt{S}\}$ similarly indicates the sensor selection (if any) for UAV $U_u\in\mathcal{U}$ to accomplish any given task $T_t\in\mathcal{T}$. It should be clear that $\lambda_{TUS}(t,u)$ exists iff (\emph{if and only if}) $\xi(u,t)=1$, i.e. UAV $U_u$ is qualified to undertake task $T_t$.

The quality of a given mission plan $\bm{\Phi}$ is driven by the Pareto optimality of a set of conflicting objectives to be simultaneously optimized:
\begin{enumerate}[noitemsep,leftmargin=*]
	\item The total cost of the mission $C^T(\bm{\Phi})$, given by $C^T(\bm{\Phi}) = \sum_{u=1}^\texttt{U} C_u T_u(\bm{\Phi})$, where $T_u(\bm{\Phi})$ is the total flight time of UAV $u$ along its routes in plan $\bm{\Phi}$.
	
	\item The end time of the mission or makespan $T^{max}(\bm{\Phi})$, i.e. the time at which the last UAV $u\in\mathcal{U}$ returns to the base.
	
	\item The risk of the mission $\delta(\bm{\Phi})\in\mathtt{R}[0,100]$, which is computed as an average percentage indicating how risky the mission is. We consider four risk factors in this problem: \gls*{uav2} that finish the mission with \emph{low fuel} (with respect to a given threshold $\mbox{FUEL}^{th}$), \gls*{uav2} that \emph{fly near to the ground} (depending on the route and the altitude $A_f(\tau)$ of the adopted flight profile through the choice of $\lambda_{TUF}$), \gls*{uav2} that \emph{fly out of the coverage or \gls*{los} of the \gls*{gcs}} controlling them; and \gls*{uav2} that \emph{fly close between them}, which intuitively depends on the time constraints between concurrently performed tasks and eventual spatial overlaps among routes/flight profiles. The risk considered in the mission relates to safety by means of factors whose values are enforced to lie within an admissible numerical range that does not interfere with the performance of the mission.	
	\item The number of \gls*{uav2} $\texttt{U}^{eff}\leq \texttt{U}$ used in the mission plan, given by $\texttt{U}^{eff}(\bm{\Phi})\doteq\sum_{u=1}^\texttt{U} \mathbb{I}(T_u(\bm{\Phi})>0)$, where $\mathbb{I}(\cdot)$ is a binary function taking value $1$ if its argument holds (and $0$ otherwise).
	
	\item The total fuel consumption $\mbox{FUEL}(\bm{\Phi})$, given by the difference between the initial fuel load $\sum_{u=1}^\texttt{U} \mbox{FUEL}_u(0)$ in the UAV prior to the start of the mission and that remaining $\sum_{u=1}^\texttt{U} \mbox{FUEL}_u(T^{max}(\bm{\Phi}))$ once the mission is over.
	
	\item The total flight time $T(\bm{\Phi})$, given by $\sum_{u=1}^\texttt{U} T_u(\bm{\Phi})$.
		
	\item The total distance $D(\bm{\Phi})$ traversed, given by $\sum_{u=1}^\texttt{U} D_u(\bm{\Phi})$, where $D_u(\bm{\Phi})$ is the total distance traversed by UAV $u$ along its routes in mission plan $\bm{\Phi}$.
\end{enumerate}

Before proceeding with the formal problem statement, some intuition must be given on the confluence of the above criteria in a many-objective design problem. On one hand, using many UAV is beneficial for minimizing the total distance traversed, the makespan of the mission, yet incurs a penalty in terms of cost. On the other hand, opting for a few UAV would yield reduced costs for the mission, but at the cost of increasing the risk of the mission. Likewise, the interplay between a moderate number of selected UAV with the above fitness measures is unclear and stringently subject to a proper choice of the number of sensors, the order in which tasks are performed and the number of UAV assigned to the task at hand, among many other decisions. Therefore, it is crucial to pose all these metrics in a many-objective optimization problem so as to assess their Pareto-optimal interactions.

The \gls*{mcmpp} problem can be then formulated as follows:
\begin{align}
& \underset{\bm{\Phi}}{\text{Min}}
& & C^T(\bm{\Phi}), T^{max}(\bm{\Phi}), \delta(\bm{\Phi}), \texttt{U}^{eff}(\bm{\Phi}), \mbox{FUEL}(\bm{\Phi}), T(\bm{\Phi}), D(\bm{\Phi}), \label{eq:minimize}\\
& \text{s. t.}
& & \lambda_{TUS}(t,u)=s \mbox{ iff } \xi(u,t)=1 \mbox{ and } \lambda_{TU}(t)=u, \label{eq:sensor}\\
&&& \sum_{u=1}^\texttt{U} \mathbb{I}(\lambda_{UG}(u)=g)\leq U_{g}^{max}, \:\forall g \in \mathcal{G}, \label{eq:maxuavgcs}\\
&&& ||\scriptsize{Pos}_u(\tau)-\scriptsize{Pos}_g||_F<R_g^{max} \mbox{ if } \lambda_{UG}(u)=g, \:\forall \tau\in[0,T^{max}(\bm{\Phi})],\label{eq:rangegcs}\\
&&& f\mbox{ : } \lambda_{TUF}(t,u)=f \mbox{ if } \lambda_{TU}(t)=u \mbox{ are feasible},\label{eq:path}\\
&&& A_f(\tau)\geq 0 \:\forall \tau \mbox{ and }\forall f\in \mathcal{F} \mbox{ : }\lambda_{TUF}(t,u)=f \mbox{ if } \lambda_{TU}(t)=u, \label{eq:minalt}\\
&&& ||\scriptsize{Pos}_{u_1}(\tau)-\scriptsize{Pos}_{u_2}(\tau)||_F>0, \:\forall u_1 \neq u_2 \in \mathcal{U}, \: \forall \tau \in [0,T^{max}(\bm{\Phi})],\label{eq:mindistuavs}\\
&&& \mbox{FUEL}_u(\tau)>0, \: \forall \tau \in[0,T^{max}(\bm{\Phi})] \mbox{ if } \lambda_{TU}(t)=u \mbox{ for some }t\in\mathcal{T}, \label{eq:fuel}\\
&&& T_u(\bm{\Phi}) \leq T_u^{max}, \: \forall u \in \mathcal{U}, \label{eq:flighttime}\\
&&& D_u(\bm{\Phi}) \leq D_u^{max}, \: \forall u \in \mathcal{U}, \label{eq:distance}\\
&&& \bm{\Phi}\mbox{ (through }\lambda_{TU}(\cdot) \mbox{ and } \Gamma_{TU} \mbox{) fulfills time and vehicle constraints} \label{eq:dependencies}
\end{align}
where Expression (\ref{eq:sensor}) denotes sensor constraints, i.e. a UAV assigned to a task should have the sensors needed to perform it installed on board; inequality (\ref{eq:maxuavgcs}) constraints the maximum number of UAV allowed by a GCS; the inequality (\ref{eq:rangegcs}) stands for a guaranteed coverage of every UAV to the GCS to which it is allocated; the notion of feasibility in Expression (\ref{eq:path}) ensure that the \gls*{uav2} perform their paths for every task and in its return to the base without any incident (as will be later detailed, a Theta* algorithm \cite{Daniel2010} is used to assure that the vehicle avoids the \gls*{nfz}, and that every climb/descent needed for the change of altitude is attainable); a positive distance to the ground, between different UAV and fuel loaded at the UAV at every time $\tau$ are reflected in inequalities (\ref{eq:minalt}), (\ref{eq:mindistuavs}) and (\ref{eq:fuel}), respectively; Expressions (\ref{eq:flighttime}) and (\ref{eq:distance}) assure that the flight time and distance traversed, respectively, for each UAV do not overpass their maximum flight time and range; and time and vehicle dependency constraints are accounted in Expression (\ref{eq:dependencies}).

\subsection{CSP Modeling of the Mission Planning Problem}\label{cspmodel}

One alternative to model the above \gls*{mcmpp} problem is by resorting to \gls*{csp}, since we seek the optimal schedule of resource-task assignments satisfying a set of imposed constraints. Thereby, we define:
\begin{itemize}[noitemsep,leftmargin=*]
\item A set of variables $\bm{\mathcal{X}} = \{x_{1},\ldots, x_{L}\}$.
\item For each variable, a finite set $\mathcal{D}_l$ of possible values (its domain).
\item And a set of constraints $\bm{\mathcal{C}} = \{C_{1}, \ldots, C_{K}\}$ restricting the values that variables can simultaneously take.
\end{itemize}

A solution to a \gls*{csp} is an assignment of a value $d_{l} \in D_{l}$ to $x_{l}\in\bm{\mathcal{X}}$ that satisfies all the imposed constraints. If all the variables have a value, then we deal with a complete instantiation of the variables. Otherwise, the variables are \emph{partially} instantiated. If we assign a value to an uninstantiated variable in a partial instantiation, then we \emph{extend} the partial instantiation. We will refer to an instantiation of a variable as $<x, v>$. The projection of a constraint $C_k$ on the variables $\{x_{1}, \ldots, x_{L'}\}$, denoted as $P(C_k, \{x_{1},\ldots, x_{L'}\})$, is the set of all $\{v_{1},\ldots, v_{L'}\}$ values for which it holds that the partial instantiation $<x_{1},v_{1}>,\ldots,<x_{L'}, v_{L'}>$ can be extended to hold the constraints. When we add time to the problem, then we have a \gls*{tcsp}. It is a particular class of \gls*{csp} where variables represent times (time points, time intervals or durations) and constraints establish allowed temporal relations between them \cite{Schwalb1998}.

The most widely used representation of a \gls*{csp} is a graph where the pairs $<$Variable,Value$>$ are the nodes and the constraints are the edges. The main techniques to solve \gls*{csp} are:

\begin{itemize}[noitemsep,leftmargin=*]
\item Search algorithms: cross the space of partial instantiations, building a complete instantiation that satisfies all the constraints, or determine that the problem is inconsistent.
\item Propagation algorithms: infer new constraints that are added to the problem without loosing any information. The worst case time complexity is polynomial in the size of the problem. They are fast in discovering local inconsistencies.
\item Structure-driven algorithms: is a mix of the two above techniques. 
\end{itemize}

\begin{figure}[!h]
		\includegraphics[width=0.95\textwidth]{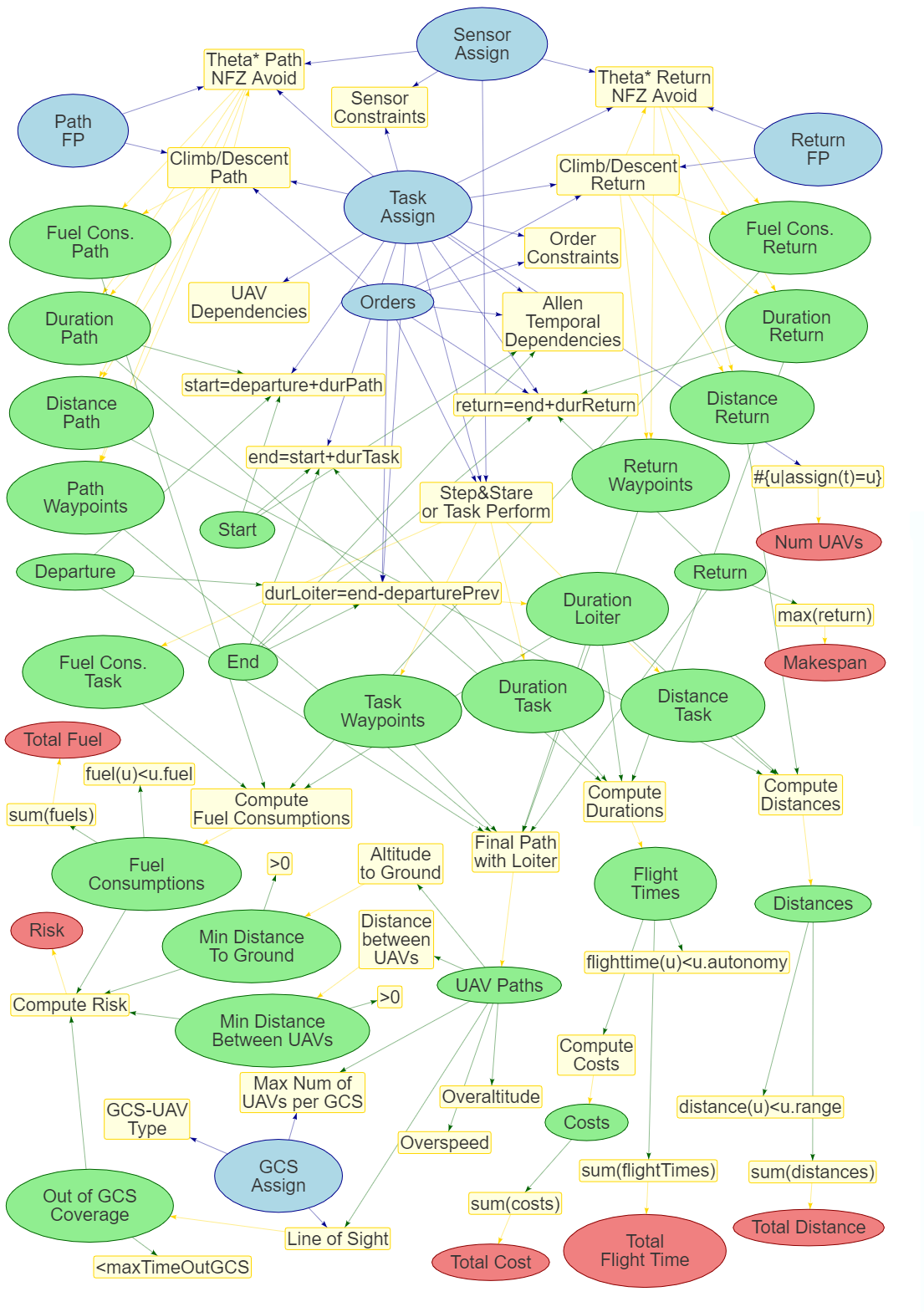}
		\centering
		\caption{Graph representation of the CSP model. Variables of the CSP are represented in blue, constraints in yellow, auxiliary variables in green and objectives in red.}
		\label{fig:cspgraph}
	\end{figure}

The \gls*{mcmpp} is modelled as a \gls*{csp}, where a graph representation of the problem is presented in Figure \ref{fig:cspgraph}. For this problem, two types of variable are used: the variables of the mission plan (represented in blue in the graph), given by the previous definition of $\bm{\Phi}$; and several auxiliary variables (represented in green in the graph) that are computed during the \gls{csp} propagation process and employed to ease the formulation of the constraints of the problem, especially those referred to time points and intervals. The different constraints of the problem are represented in yellow boxes, and the optimization objectives are represented in red ovals.

The variables of the \gls*{csp} are then:
\begin{itemize}[noitemsep,leftmargin=*]
\item \emph{Task Assign}: this variable corresponds to $\lambda_{TU}$ in the \gls*{mcmpp}.
\item \emph{Orders}: this variable corresponds to $\Gamma_{TU}$ in the \gls*{mcmpp}.
\item \emph{GCS Assign}: this variable corresponds to $\lambda_{UG}$ in the \gls*{mcmpp}.
\item \emph{Sensor Assign}: this variable corresponds to $\lambda_{TUS}$ in the \gls*{mcmpp}.
\item \emph{Path FP}: this variable corresponds to the first $\texttt{T} \cdot \texttt{U}$ variables $\{1,\ldots,\texttt{T}\}\times\{1,\ldots,\texttt{U}\}$ of $\lambda_{TUF}$ in the \gls*{mcmpp}, i.e. it represents the flight profiles used in all the paths perform by \gls*{uav2} for going to tasks, but not the return path.
\item \emph{Return FP}: this variable corresponds to the last $\texttt{U}$ variables of $\lambda_{TUF}$ in the \gls*{mcmpp}, i.e. it represents the flight profiles used in the $\texttt{ReturnToBase}$ paths (i.e. the $(\texttt{T}+1)$-th task) for every \gls*{uav2} used in the mission.
\end{itemize}

On the other hand, the auxiliary variables used in the \gls*{csp} model are:
\begin{itemize}[noitemsep,leftmargin=*]
\item \emph{Departure}: This variable represents the time point for each UAV/task combination when the \gls*{uav2} departs from its position to go to the task zone.
\item \emph{Start}: This variable represents the time point for each UAV/task combination when the \gls*{uav2} arrives to the task zone and starts the performance of the task.
\item \emph{End}: This variable represents the time point for each UAV/task combination when the \gls*{uav2} finishes the task performance.
\item \emph{Return}: This variable represents the time point for each UAV when it has returned and landed into its base position, and consequently finished its duties in the mission.
\item \emph{Duration Path}, \emph{Distance Path}, \emph{Fuel Cons. Path} and \emph{Path Waypoints}: These variables represent, respectively, the time elapsed, the distance traversed, the fuel consumed and the waypoints followed between the departure and the start of each UAV/task combination.
\item \emph{Duration Task}, \emph{Distance Task}, \emph{Fuel Cons. Task} and \emph{Task Waypoints}: These variables represent, respectively, the time elapsed, the distance traversed, the fuel consumed and the waypoints followed between the start and the end of each UAV/task combination.
\item \emph{Duration Return}, \emph{Distance Return}, \emph{Fuel Cons. Return} and \emph{Return Waypoints}: These variables represent, respectively, the time elapsed, the distance traversed, the fuel consumed and the waypoints followed between the end of the last task performed and the return for each \gls*{uav2}.
\item \emph{Duration Loiter}: This variable represents the time elapsed for each UAV/task combination between the end of the previous task performed by the \gls*{uav2} and the departure for the present task. This variable is higher than 0 when there exists some time dependencies that impede the immediate performance of the task.
\item \emph{UAV Paths}: This variable represents the waypoints followed by each \gls*{uav2} for the whole mission.
\item \emph{Fuel Consumptions}: This variable represents the fuel consumed by each \gls*{uav2} in the whole mission, i.e. $\mbox{FUEL}_u(0)-\mbox{FUEL}_u(T^{max})$.
\item \emph{Flight Times}: This variable represents the flying time of each \gls*{uav2} during the whole mission, i.e. $T_u$.
\item \emph{Distances}: This variable represents the distance traversed by each \gls*{uav2} in the whole mission, i.e. $D_u$.
\item \emph{Costs}: This variable represents the cost spent for each \gls*{uav2} in the mission, i.e. $C_u T_u$.
\item \emph{Min Distance To Ground}: This variable represents the minimum altitude of each \gls*{uav2} to the ground (given by an elevation file) during its mission flight, i.e. $min_{\tau \in [T_u^{Climb},T_u^{Descent}]} A_u(\tau)$, where $T_u^{Climb}$ and $T_u^{Descent}$ represent, respectively, the climb point when the UAV has reached the flight profile altitude after the take-off, and the descent point when the UAV starts the landing.
\item \emph{Min Distance Between UAV}: This variable represents the minimum distance between each pair of \gls*{uav2} in the mission (lowest distance in all temporal points), i.e. $min_{\tau \in [0,T^{max}]} ||\scriptsize{Pos}_{u_1}(\tau)-\scriptsize{Pos}_{u_2}(\tau)||_F$.
\item \emph{Out of GCS Coverage}: This variable represents the accumulated time for each UAV where \gls*{los} is not maintain with the \gls*{gcs} in the mission.
\end{itemize}

Finally, the constraints of the \gls*{csp} considered are:
\begin{itemize}[noitemsep,leftmargin=*]
\item \emph{Sensor constraints}: they check if a \gls*{uav2} has the sensors needed to perform its assigned tasks. They are equivalent to Equation (\ref{eq:sensor}) in the \gls*{mcmpp} definition.
\item \emph{Order constraints}: they check that the orders for two different tasks assigned to the same \gls*{uav2} are different, and lower than the number of tasks assigned to that \gls*{uav2}, i.e. $\Gamma_{TU}(t_1,u) \neq \Gamma_{TU}(t_2,u), \: \forall t_1 \neq t_2 \in \mathcal{T} \: \forall u \in \mathcal{U}$.
\item \emph{Allen Temporal Dependencies} and \emph{UAV Dependencies}: these constraints are related to the time and vehicle dependencies mentioned previously and represented in Equation (\ref{eq:dependencies}) in the \gls*{mcmpp} definition.
\item \emph{GCS constraints}: they assure that the \gls*{gcs} assignments are correct, \gls*{uav2} are assigned to \gls*{gcs} able to control them (\emph{GCS-UAV Type}); \gls*{gcs} do not overpass their maximum number of \gls*{uav2}(\emph{Max Num of UAV per GCS}) expressed in Equation (\ref{eq:maxuavgcs}); and they are located within the \gls*{gcs} coverage area (\emph{Line of Sight}), expressed in Equation (\ref{eq:rangegcs}).
\item \emph{Path constraints}: these constraints assure that the \gls*{uav2} perform its path for every UAV/task combination without any incident. In this sense, a Theta* algorithm \cite{Daniel2010} is used to assure that the vehicle avoids the \gls*{nfz} (\emph{Theta* Path NFZ Avoid}), and that every climb/descent needed for the change of altitude (\emph{Climb/Descent Path}) is attainable. These constraints are part of the Equation (\ref{eq:path}) in the \gls*{mcmpp} definition.
\item \emph{Return constraints}: they assure that each \gls*{uav2} perform its return to the base without any incident, including \gls*{nfz} avoidance (\emph{Theta* Return NFZ Avoid}) and climb/descent routes (\emph{Climb/Descent Return}). These constraints are part of the Equation (\ref{eq:path}) in the \gls*{mcmpp} definition.
\item \emph{Step\&Stare or Task Perform}: they compute the task performance waypoints, distances, durations and fuel consumptions according to the specific task, e.g. Mapping tasks perform a Step\&Stare over the task zone.
\item \emph{Temporal constraints}: they assure the consistency of all the times and durations involved in the mission planning. This includes that $start=departure+durPath$, $end=start+durTask$, $return=end+durReturn$ and $durLoiter=departure-endPrev$.
\item \emph{Overaltitude} and \emph{Overspeed}: these constraints assure that the waypoints for each \gls*{uav2} (\emph{Final Path with Loiter}) do not overpass the speed and altitude limit of the vehicle.
\item \emph{Distance to ground $>0$}: these constraints assure that the vehicle does not collide with the terrain. This is equivalent to Equation (\ref{eq:minalt}) in the \gls*{mcmpp} definition.
\item \emph{Distance between vehicles $>0$}: they assure that \gls*{uav2} do not collide during the mission. This is equivalent to Equation (\ref{eq:mindistuavs}) in the \gls*{mcmpp} definition.
\item \emph{Fuel constraints} ($fuel(u)<u.fuel$): they assure that the fuel consumed by each vehicle is less than its initial fuel. This is equivalent to Equation (\ref{eq:fuel}) in the \gls*{mcmpp} definition.
\item \emph{Flight time constraints} ($flighttime(u)<u.maxFlightTime$): they assure that the total flight time for each vehicle is less than its vehicle maximum flight time. This is equivalent to Equation (\ref{eq:flighttime}) in the \gls*{mcmpp} definition.
\item \emph{Distance constraints} ($distance(u)<u.range$): they assure that the distance traversed by each vehicle is less than its range. This is equivalent to Equation (\ref{eq:distance}) in the \gls*{mcmpp} definition.
\end{itemize}

\section{Proposed Weighted-Random Multi-Objective Evolutionary Algorithm for Mission Planning}\label{pmogammp}
Real-world decision problems often require the solutions to meet multiple performance criteria (or objectives) simultaneously, for which they are referred to as \gls*{mop}. Such objectives are often conflicting, wherein an improvement in one objective cannot be achieved without detriment to another objective. In this case, there is no single solution to a \gls*{mop} that can be selected objectively; rather a set of solutions exists, representing different performance trade-offs between the considered criteria. A minimization \gls*{mop} can be mathematically defined as follows:
\begin{equation}
\underset{\mathbf{x}}{\text{Min}}\, \mathbf{f}(\mathbf{x})=(f_1(\mathbf{x}), f_2(\mathbf{x}), ..., f_m(\mathbf{x}))^T \quad \text{ subject to } \mathbf{x} \in \Omega \subseteq \mathbb{R}^n
\end{equation}
where $\mathbf{x}=(x_1,x_2,...,x_n)^T$ is a vector of $n$ decision variables from the decision space $\Omega$; $\mathbf{f}:\Omega \rightarrow \Theta \subseteq \mathbb{R}^m$ consists of a set of $m$ objective functions, and a mapping from the $n$-dimensional decision space $\Omega$ to the $m$-dimensional objective space $\Theta$.

\begin{definition}{}
Given two decision vectors $\mathbf{x},\mathbf{y}\in \Omega$, $\mathbf{x}$ is said to \emph{Pareto dominate} $\mathbf{y}$, denoted by $\mathbf{x} \prec \mathbf{y}$, if:
\begin{align}
\forall i \in \{1,2,...,m\} \quad f_i(\mathbf{x}) \leq f_i(\mathbf{y}) \nonumber \\
\exists j \in \{1,2,...,m\} \quad f_j(\mathbf{x}) < f_j(\mathbf{y})
\end{align}
\end{definition}

\begin{definition}{}
A decision vector $\mathbf{x}^*\in \Omega$, is \emph{Pareto optimal} if $\nexists \mathbf{x} \in \Omega$, $\mathbf{x} \prec \mathbf{x}^*$.
\end{definition}

\begin{definition}{}
The \emph{Pareto set} $PS$, is defined as:
\begin{equation}
PS=\{\mathbf{x} \in \Omega|\mathbf{x} \text{ is Pareto optimal}\}
\end{equation}
\end{definition}

\begin{definition}{}
The \emph{Pareto front} $PF$, is defined as:
\begin{equation}
PF=\{\mathbf{f}(\mathbf{x}) \in \mathbb{R}^m|\mathbf{x} \in PS\}
\end{equation}
\end{definition}

\begin{figure}[!h]
	\includegraphics[width=\textwidth]{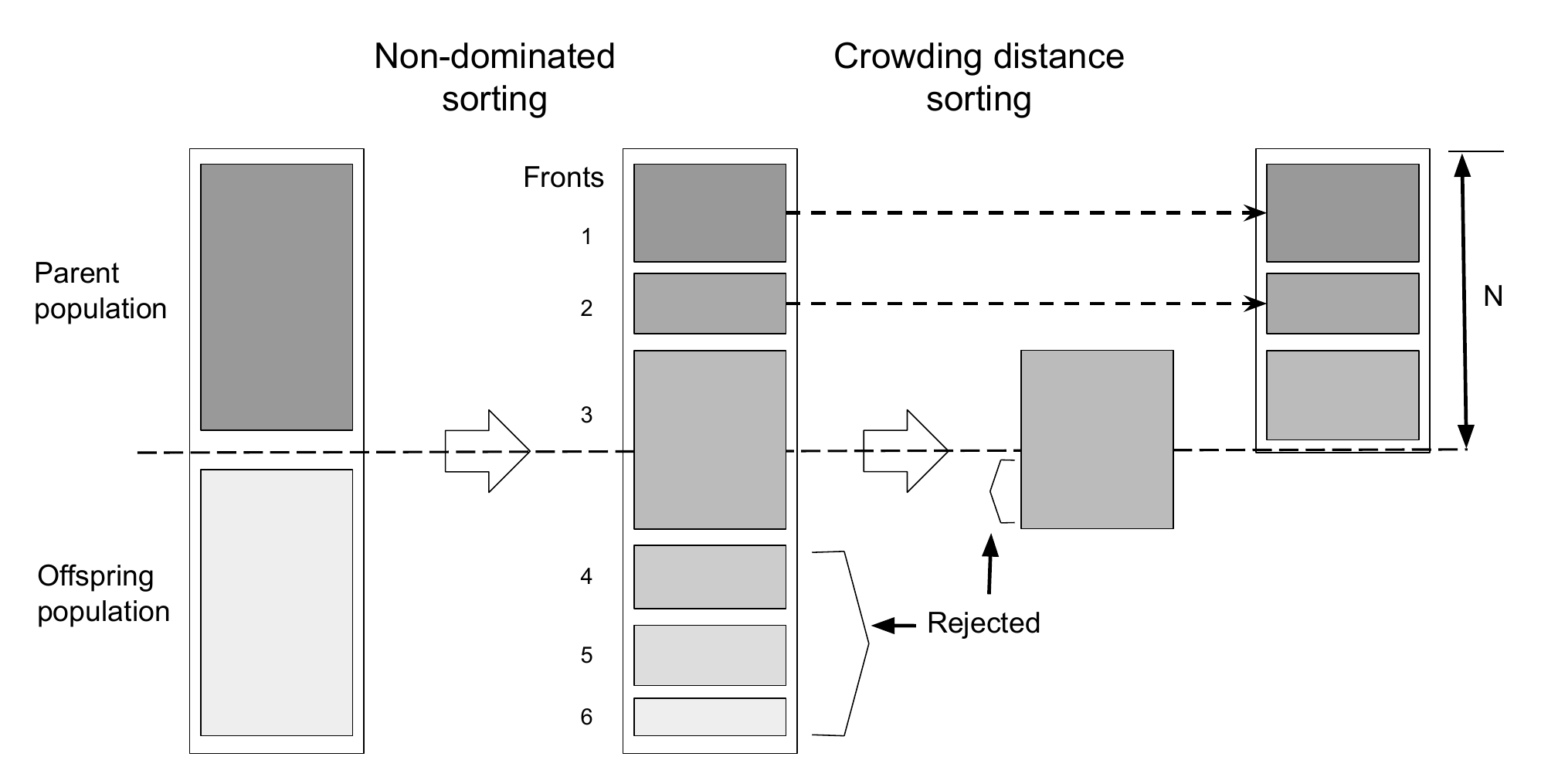}
	\centering
	\caption{NSGA-II overview of multiobjective strategies.}
	\label{fig:nsga2}
\end{figure}

The goal of \gls*{emo} algorithms is to evolve the non-dominated objective vectors towards $PF$ (convergence), and also generate a good distribution of these vectors over the $PF$ (diversity).
The most used algorithm over the last decade in this field has been the \gls*{nsga2} \cite{Deb2002}. \Gls*{nsga2} achieves convergence and diversity by relying on two measures when comparing individuals (e.g. for selection and deletion): The first is the non-domination rank, which measures how close an individual is to the non-dominated front. An individual with a lower rank (closer to the front) is always preferred to an individual with a higher rank. If two individuals have the same non-domination rank, as a secondary criterion, a crowding measure is used, which prefers individuals which are in rather deserted areas of the front. More precisely, for each individual the cuboid length is calculated, which is the sum of distances between an individual’s two closest neighbours in each dimension. The individuals with greater cuboid length are then preferred. Figure \ref{fig:nsga2} shows an overview of these strategies.

In this approach, a previous hybrid approach based on \gls*{moea} and \gls*{csp} \cite{Ramirez-Atencia2017} for the \gls*{mcmpp} is extended to consider the characteristics of the problem. The \gls*{csp} is computed inside the fitness function of the \gls*{moea}, checking that solutions fulfill all the constraints. In addition, to deal with the huge search space of the problem and guide the algorithm to find valid solutions, a biased random generator has been designed for the initializer and mutation operator.

\subsection{Encoding}
The encoding of this new approach consists of six different alleles representing the features to be assigned in the \gls*{mcmpp} (presented in Section \ref{cspmodel}), as can be seen in Figure \ref{fig:chromosome}:

	\begin{figure}[!h]
		\includegraphics[width=0.5\textwidth]{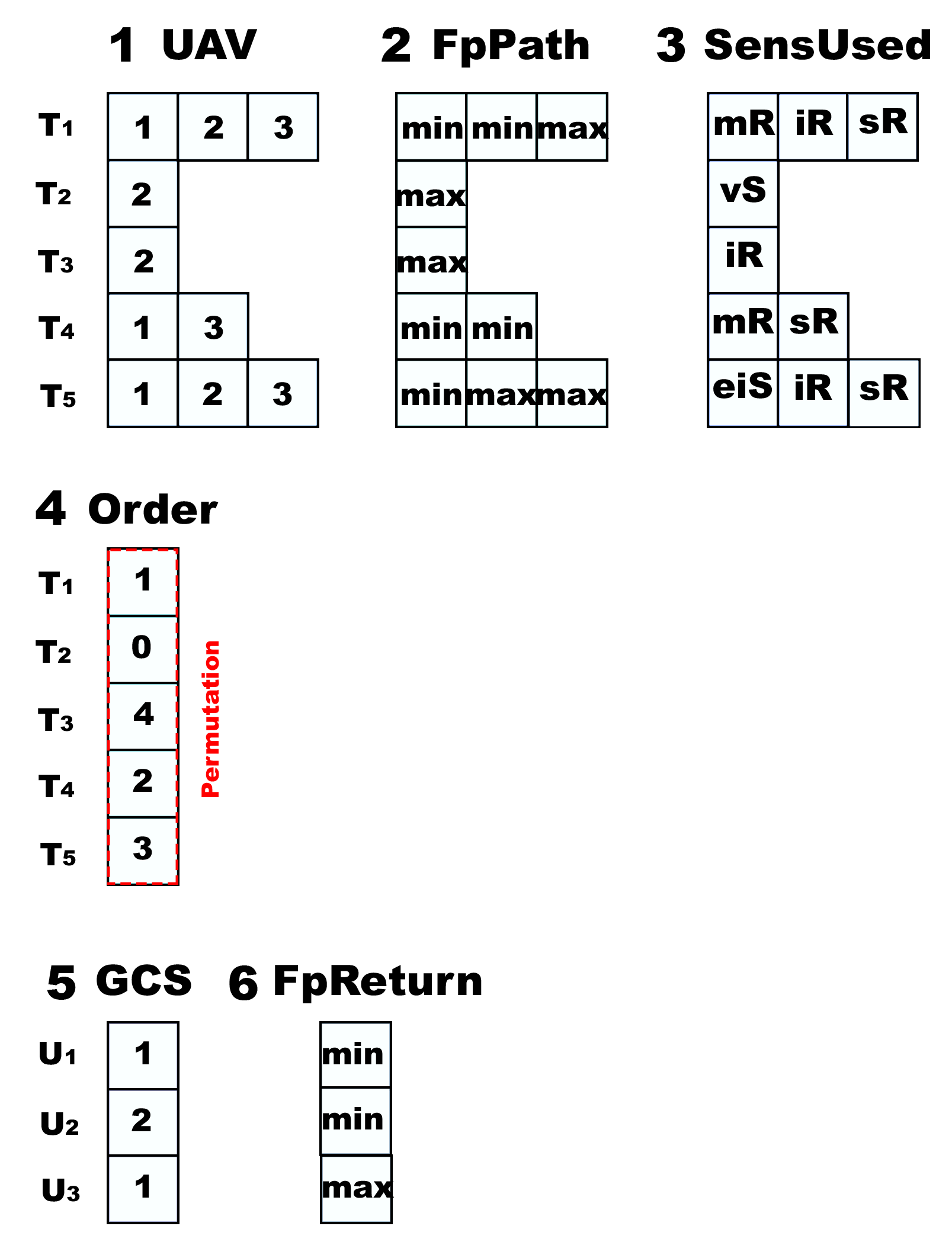}
		\centering
		\caption{Example of an individivual that represents a possible solution for a problem with 5 tasks, 3 \gls*{uav2} and 2 \gls*{gcs}.}
		\label{fig:chromosome}
	\end{figure}
	
\begin{enumerate}
	\item \emph{\gls*{uav2} assigned to each task}. This allele corresponds to the \textit{Task Assign variable} of the \gls{csp} model. If the $T_i$ task is Multi-\gls*{uav2}, then this cell contains a vector representing the different \gls*{uav2} assigned to this task, as shown in Figure \ref{fig:chromosome} for tasks $T_1$, $T_3$ and $T_4$. The number of combinations, or space size, for this allele is, considering $\texttt{T}_{MU}$ the number of Multi-UAV tasks, $\texttt{U}^{\texttt{T}-\texttt{T}_{MU}} \times (2^{\texttt{U}}-1)^{\texttt{T}_{MU}}$.
		\item \emph{Flight profiles used for each UAV  in the path to an assigned task}. This allele corresponds to the \textit{Path FP variable} of the \gls{csp} model. As in the first allele, some of the cells could contain a vector if the corresponding task is performed by several \gls*{uav2}. The possible values considered in this approach are \emph{min} (minimum consumption profile) and \emph{max} (maximum speed profile). The space size for this allele is $2^{\texttt{T}-\texttt{T}_{MU}} \times (2^{\texttt{U}+1}-2)^{\texttt{T}_{MU}}$.
	\item \emph{Sensors used for the task performance} by each \gls*{uav2}. This allele corresponds to the \textit{Sensor Assign variable} of the \gls{csp} model. These variables are necessary when the vehicle performing the task has several sensors that could perform that task. The values considered in this approach are eiS (\gls*{eoir} sensor), sR (\gls*{sar} radar), iR(\gls*{isar} radar) and mR (\gls*{mpr} radar). The space size for this allele depends on the number of sensors for each vehicle. Considering that all vehicles have $s$ sensors, then the space size is $s^{\texttt{T}-\texttt{T}_{MU}} \times (\sum_{i=1}^{\texttt{U}} s^i)^{\texttt{T}_{MU}}$.
	\item \emph{Permutation of the task orders}. This allele corresponds to the \textit{Orders variable} of the \gls{csp} model. These values indicate the absolute order of the tasks. It is only used if there are several tasks assigned to the same \gls*{uav2} (e.g. in Figure \ref{fig:chromosome}, \gls*{uav2} 1 performs tasks 1, 4 and 5 in this order). In a previous publication \cite{Ramirez-Atencia2015}, it was shown that this representation, although it is not injective, is more efficient than using the orders for each pair task-UAV. The space size for this allele is $\texttt{T}!$.
	\item \emph{\gls*{gcs} controlling each UAV}. What \gls*{gcs} is assigned to monitor what UAV. This allele corresponds to the \textit{GCS Assign variable} of the \gls{csp} model. The space size for this allele is $\texttt{G}^{\texttt{U}}$.
	\item \emph{Flight Profiles used by each} UAV to return to the base. This allele corresponds to the \textit{Return FP variable} of the \gls{csp} model. The space size for this allele is $2^{\texttt{U}}$.
\end{enumerate}

The multiplication of the space sizes for each allele provides the total search space size of the problem, which shows the high complexity of adding a task, \gls*{uav2} or \gls*{gcs} to the problem.


In order to reduce this huge search space, some constraints have been used within the initialization and the mutation operator in order to assure that the solutions generated are more likely to be valid. These constraints considered are:

\begin{itemize}[noitemsep,leftmargin=*]
\item \emph{Sensor constraints}: In the task assignments, a \gls*{uav2} is selected if and only if it has the appropriate sensor to perform the task.
\item \emph{\gls*{gcs}-\gls*{uav2} type constraints}: In the \gls*{gcs} assignments, a \gls*{gcs} is selected for controlling a \gls*{uav2} if and only if the \gls*{gcs} is able to handle that type of vehicle.
\item \emph{Allen Temporal Dependencies}: In the order permutation, when generating the permutation, if there is some precedence or \emph{meets} dependency, then this precedence is checked, and if not satisfied, these values are swapped in the permutation.
\end{itemize}

\subsection{Weighted strategies for biased random generator}
Usually, the first population for a \gls*{moea} is generated randomly. But sometimes, in order to reduce the convergence time, if some knowledge about the problem is provided, it can be used to guide the search of solutions.

This information can be expressed as a high probability but not absolute certainty of (e.g. in the \gls*{vrp} with multiple vehicles \cite{Iqbal2015,VenkataNarasimha2013}, it is highly probable that the best vehicle for going to a waypoint is the nearest to it in the space; but this is not always true as sometimes it implies other waypoints far from this but also near to the vehicle to remain unassigned due to fuel or flight time constraints). In order to consider this kind of information appropriately, weighted strategies can be used. These strategies are used instead of the typical random generator inside the \gls*{moea}, in both creation and mutation, and assigns different probabilities for each of the possible values of the genes.

In this paper, three strategies are proposed to be considered:

\begin{itemize}[noitemsep,leftmargin=*]
\item Arithmetic strategy: This strategy gives a lower probability to less probable values following an inverse arithmetic progression (i.e. $\frac{N - i}{N}$, with $N$ the biggest value and $i$ the value considered). For an integer set of $N$ values where the bigger the value, the lower its probability, the weight for each value using the arithmetic strategy will be as follows:
\begin{equation}
\forall i \in 1..N, \quad weight[i]=\frac{N - i}{N}
\end{equation}

\item Harmonic strategy: This strategy gives a lower probability to less probable values following a harmonic progression (i.e. $\frac{1}{i}$, with $i$ the value considered). For an integer set of $N$ values where the bigger the value, the lower its probability, the weight for each value using the harmonic strategy will be as follows:
\begin{equation}
\forall i \in 1..N, weight[i]=\frac{1}{i}
\end{equation}

\item Geometric strategy: This strategy gives a lower probability to less probable values following an inverse geometric progression (i.e. $\frac{1}{2^{i}}$, with $i$ the value considered). For an integer set of $N$ values where the bigger the value, the lower its probability, the weight for each value using the geometric strategy will be as follows:
\begin{equation}
\forall i \in 1..N, weight[i]=\frac{1}{2^{i}}
\end{equation}
\end{itemize}

\begin{figure}[!h]
		\includegraphics[width=\textwidth]{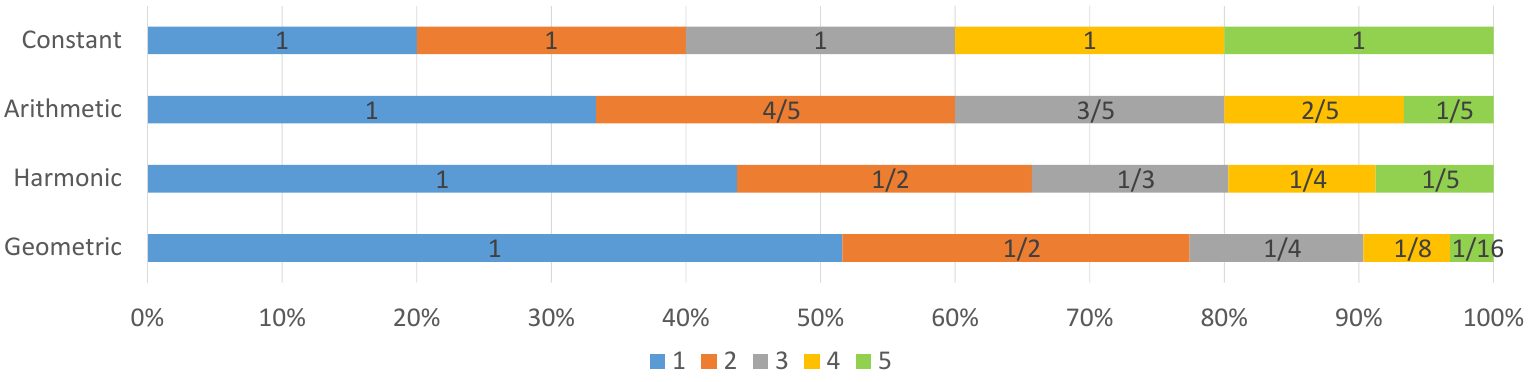}
		\centering
		\caption{Weighted strategies considered with 5 integer values of decreasing probability.}
		\label{fig:strategies}
	\end{figure}

These different strategies are represented in Figure \ref{fig:strategies} for a case with a maximum of 5 integer values with decreasing probability. In this figure, the constant strategy represents the typical random generator, where all values have the same probability.

Here, we can see that with the geometric approach, it is more likely to select value 1, while the harmonic and the arithmetic approach have lower probabilities. With the arithmetic approach, the decrease in the probabilities is linear, while in the geometric approach the decay is exponential. On the other hand, the harmonic approach provides a larger decrease for the first values, but lower for the last ones. Based on this, the use of the arithmetic approach is more suitable when the certainty of selecting the most likely value is not very high, so other less probable values could be easily selected. The harmonic approach should be instead used when the certainty of selecting the most probable value is high and selecting the less probable values is indifferent for the final result. Finally, the geometric approach becomes useful when the certainty of selecting the most probable value is very high, and the less certain any other value is, the less likely is to obtain good results using this weight distribution.

In this approach, we have implemented a biased random generator for the formation of new individuals. This biased random generator is applied in three parts of the encoding:

\begin{itemize}[noitemsep,leftmargin=*]
\item \emph{Number of UAV Selection (NUS)}: for Multi-UAV tasks, the selection of the number of \gls*{uav2} is performed using a weighted strategy. In the previous hybrid approach, most of the optimization objectives (except for the makespan) used as fitness function, obtained better results when the number of \gls*{uav2} assigned for these tasks is low. For this reason, the new weighted random function provides a lower probability for higher numbers of \gls*{uav2}.

\item \emph{Distance-based UAV Selection (DUS)}: for any task, the selection of the \gls*{uav2}(s) to perform it, is done through a weighted strategy dependant on the distance from the \gls*{uav2}(s) to the task. In each mission, there are different vehicles that can perform the same task. Due to this, the selection of the \gls*{uav2}(s) commanded to perform a specific task could hinge on the distance of every \gls*{uav2}(s) to the task at hand. \Gls*{uav2}(s) closer to the task would spent less time to arrive to the location where the task must be performed, and therefore obtain better solutions with less fuel consumption, flight time, etc. So, the closer the \gls*{uav2} is to the task, the higher the probability of being selected.

\item \emph{Distance-based GCS Selection (DGS)}: for any UAV, the selection of the \gls*{gcs} controlling it, is done through a weighted strategy dependant on the distance from the \gls*{uav2} to the \gls*{gcs}. The closer the \gls*{gcs} is to the \gls*{uav2}, the higher the probability of being selected.
\end{itemize}

In order to assume different orders of consistency for the selection of the \gls*{uav2} number, several strategies have been proposed for the NUS weighted random function:

\begin{itemize}[noitemsep,leftmargin=*]
\item Arithmetic: $\forall i=1..\texttt{U}, weight[i]=\frac{\texttt{U} - i}{\texttt{U}}$
\item Harmonic: $\forall i=1..\texttt{U}, weight[i]=\frac{1}{i}$
\item Geometric: $\forall i=1..\texttt{U}, weight[i]=\frac{1}{2^{i}}$
\end{itemize}

These different strategies are represented in Figure \ref{fig:strategies} for a case with a maximum of 5 \gls*{uav2}. Here, we can see that with the geometric approach, it is more likely to select just 1 \gls*{uav2}, while the harmonic and the arithmetic approach have lower probabilities. The constant approach (uniform random function) gives the same probability to every possible number of \gls*{uav2} to be used.

On the other hand, the same strategies have been implemented for the DUS weighted random function. However, in this case instead of the number of \gls*{uav2}, the distance from each one to the task is used:

\begin{itemize}[noitemsep,leftmargin=*]
\item Arithmetic: $\forall t \in \mathcal{T}, \forall u \in \mathcal{U}, weight[t,u]=\frac{\max_{j \in  \mathcal{U}} ||\scriptsize{Pos}_u(0)-\scriptsize{Pos}_j||_F - ||\scriptsize{Pos}_u(0)-\scriptsize{Pos}_t||_F}{\max_{j \in  \mathcal{U}} ||\scriptsize{Pos}_u(0)-\scriptsize{Pos}_j||_F}$
\item Harmonic: $\forall t \in \mathcal{T}, \forall u \in \mathcal{U}, weight[t,u]=\frac{1}{||\scriptsize{Pos}_u(0)-\scriptsize{Pos}_t||_F}$
\item Geometric: $\forall t \in \mathcal{T}, \forall u \in \mathcal{U}, weight[t,u]=\frac{1}{2^{||\scriptsize{Pos}_u(0)-\scriptsize{Pos}_t||_F}}$
\end{itemize}

Finally, in a similar way, these strategies are used with the DGS weighted random function:

\begin{itemize}[noitemsep,leftmargin=*]
\item Arithmetic: $\forall u \in \mathcal{U}, \forall g \in \mathcal{G}, weight[u,g]=\frac{\max_{j \in \mathcal{G}} ||\scriptsize{Pos}_u(0)-\scriptsize{Pos}_j||_F - ||\scriptsize{Pos}_u(0)-\scriptsize{Pos}_g||_F}{\max_{j \in \mathcal{G}} ||\scriptsize{Pos}_u(0)-\scriptsize{Pos}_j||_F}$
\item Harmonic: $\forall u \in \mathcal{U}, \forall g \in \mathcal{G}, weight[u,g]=\frac{1}{||\scriptsize{Pos}_u(0)-\scriptsize{Pos}_g||_F}$
\item Geometric: $\forall u \in \mathcal{U}, \forall g \in \mathcal{G}, weight[u,g]=\frac{1}{2^{||\scriptsize{Pos}_u(0)-\scriptsize{Pos}_g||_F}}$
\end{itemize}

\subsection{Algorithm}
\begin{figure}[!h]
		\includegraphics[width=0.8\textwidth]{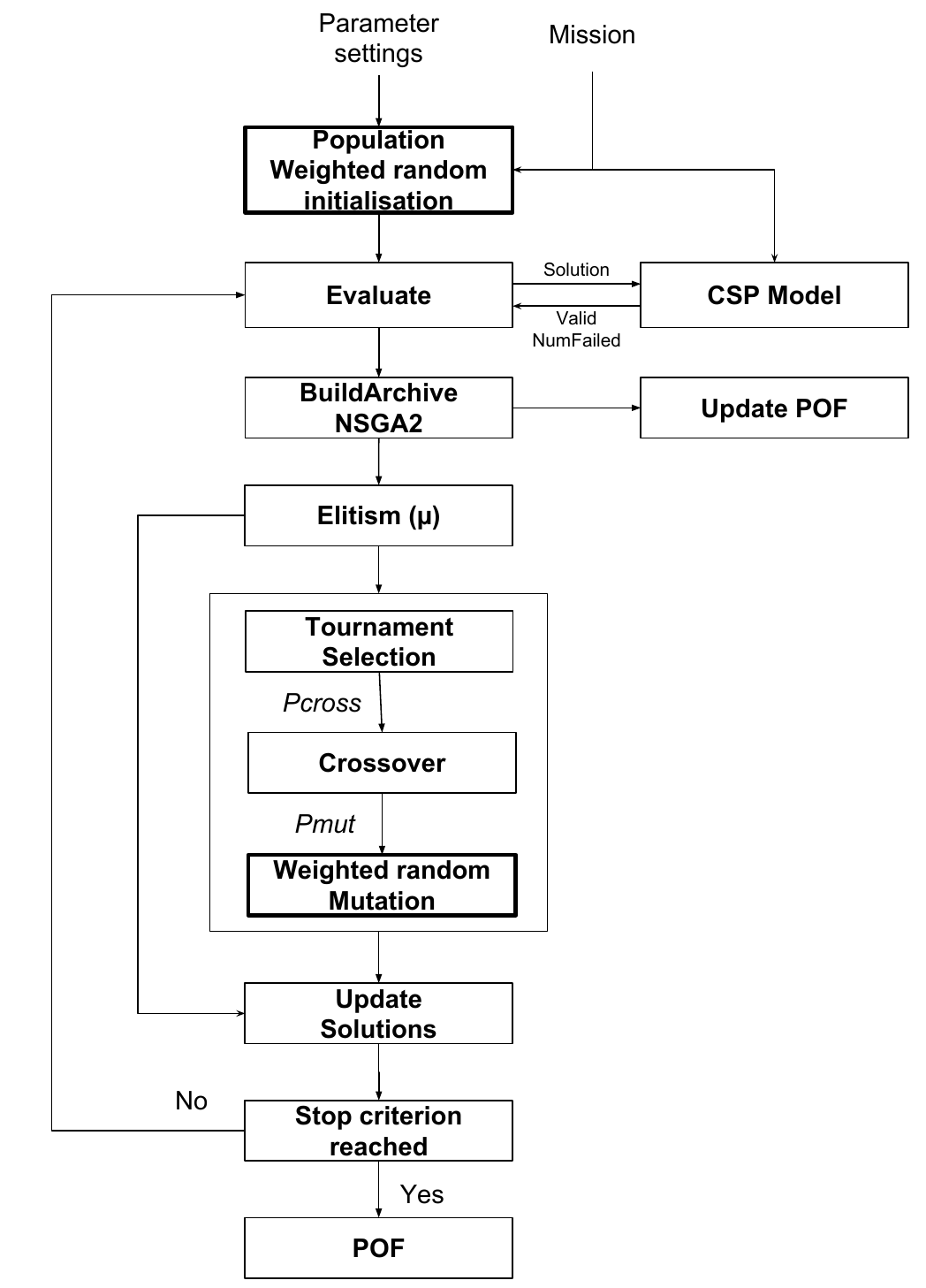}
		\centering
		\caption{Overview of WRNSGA-II algorithm for mission planning.}
		\label{fig:wrnsga2}
	\end{figure}

The \gls*{wrnsga2} is presented in Figure \ref{fig:wrnsga2}. In this new approach, firstly a \emph{weighted random generation of the initial population} is performed. Then this initial population evolves using a multi-objective evolutionary algorithm based on \gls*{nsga2} approach \cite{Deb2002}.

The population weighted random initialisation generates a set of $\lambda$ weighted random individuals, each one generated using the function presented in Algorithm \ref{alg:weightedinit}. First, the number of \gls*{uav2} to be used when tasks are Multi-UAV is selected (Lines 4-8) using the provided NUS strategy. Then, the concrete \gls*{uav2} to be used are selected (Lines 10-14) according to the provided DUS strategy. After that, the permutation orders, path flight profiles and sensors used are selected randomly as usual. In Lines 20-23, it is shown how the \gls*{gcs} controlling the vehicles are selected according to the DGS strategy. In this algorithm, the costs computed for strategies are represented as $cNUS$, $cDUS$ and $cDGS$. In both DUS and DGS, the geodesic distance function used to compute the distances between tasks, vehicles and ground stations is represented as $d$.

\begin{algorithm}[!h]
 \caption{WeightedRandomIndividual($M$,$NUS$,$DUS$,$DGS$)}
 \label{alg:weightedinit}
\DontPrintSemicolon
\KwIn{ A mission $M=(\mathcal{T},\mathcal{U},\mathcal{G})$ where $\mathcal{T}$ is a set of tasks to perform denoted by $\{t_1, \dots, t_\texttt{T}\}$, $\mathcal{U}$ is a set of UAV denoted by $\{u_1, \dots, u_\texttt{U}\}$ and $\mathcal{G}$ is a set of GCSs denoted by $\{g_1, \dots, g_\texttt{G}\}$. And weighted strategies for $NUS$, $DUS$ and $DGS$}
\KwOut{ A Weighted random individual}
 $ind \gets \emptyset$\;
 \For{$t_i \in \mathcal{T}$}{
   $num_{UAV} \gets 1$ \;
   \If {isMultiUAV($t_i$)}{
	  $cNUS \gets [Strategy(j,\texttt{U},NUS)]_{j=1}^{\texttt{U}}$\;
      $num_{UAV} \gets weightedRandomValue($[1,2,...,\texttt{U}]$,cNUS)$ \;
   }
   $t_{assign} \gets \emptyset$ \;
   \While{$|t_{assign}| < num_{UAV}$}{
	  $cDUS \gets [Strategy(d(t_i,u_j),\max_{u_k \in \mathcal{U}}{d(t_i,u_k)},DUS)]_{u_j \in \mathcal{U}}$\;
      $t_{assign} \gets t_{assign} \cup weightedRandomValue(\mathcal{U},cDUS)$ \;
   }
   $ind.UAV[t_i] \gets t_{assign}$\;
 }
 $assignPermutationOrders(ind,M)$\;
 $assignFPPaths(ind,M)$\;
 $assignSensors(ind,M)$\;
 \For{$u_i \in \mathcal{U}$}{
   $cDGS \gets [Strategy(d(u_i,g_j),\max_{g_k \in \mathcal{G}}{d(u_i,g_k)},DGS)]_{g_j \in \mathcal{G}}$\;
   $ind.GCS[u_i] \gets weightedRandomValue(\mathcal{G},cDGS)$ \;
 }
 $assignFPReturns(ind,M)$\;
 \Return{ind}\;
\end{algorithm}

For all these selections, a general strategy function returning the cost for each value has been designed (Algorithm \ref{alg:strategy}), as well as a weighted random function which returns a weighted random value according to the costs provided (Algorithm \ref{alg:weightedvalue}).

\begin{algorithm}[!h]
 \caption{Strategy($value$,$max$,$strategy$)}
 \label{alg:strategy}
\DontPrintSemicolon
\KwIn{ A $value$ to which the concrete $strategy$ must be applied, and the maximum value $max$ between all possible values. }
\KwOut{The concrete cost for the value}
 \Switch{$strategy$}{
 	\Case{Arithmetic}{
 		\Return{$\frac{max-value}{max}$}\;
 	}
   \Case{Harmonic}{
		\Return{$\frac{1}{value}$}\;
   }
   \Case{Geometric}{
		\Return{$\frac{1}{2^{value}}$}\;
   }
   \Case{Constant}{
		\Return{$1$}\;
   }
 }
\end{algorithm}

\begin{algorithm}[!h]
 \caption{weightedRandomValue(values,costs)}
 \label{alg:weightedvalue}
\DontPrintSemicolon
\KwIn{ Vectors $V$ (integer values considered) and $C$ (double costs for each value) of size $d$}
\KwOut{A weighted random value based on the costs}
 $total \gets$ sum($C$) \;
 $randomValue \gets randomDouble() \cdot total$ \;
 \For{$i \gets 1$ \textbf{to} $d$}{
 	$randomValue \gets randomValue - C_i$\;
   \If {$randomValue \leq 0$}{
   	 \Return{$V_i$} \;
   }
 }
\end{algorithm}

The evaluation of the individuals is performed by a fitness function that first checks that all the constraints of the \gls*{mcmpp} are fulfilled for a given solution. For this, a \gls*{csp} solver is used to check if the constraints are satisfied for a concrete solution. When the solution is not valid according to the \gls{csp}, then the number of unfulfilled constraints is returned, and used as the optimization criteria with other invalid solutions. On the other hand, if all constraints are fulfilled, the fitness works as a multi-objective function minimizing the objectives of the problem presented in Section \ref{formulation}.

A tournament selection is used to provide the individuals that will be chosen to apply the genetic operators. The crossover operator consists of an extension of the 2-point crossover and the \gls*{pmx} independently applied to each allele of the chromosome, where a 2-point crossover is applied to the 3 first alleles of size $\texttt{T} \cdot \texttt{U}$, \gls*{pmx} is applied to the fourth allele (the permutation of orders), and another 2-point crossover is applied to the fifth and sixth alleles of size $\texttt{U}$. The mutation operator is also an extension of the uniform and the insert mutation applied to each allele using the same criterion as in the crossover. This \emph{mutation operator} is \emph{combined} with the \emph{biased random generator} in order to produce new individuals based on the weighted random functions previously explained.

Finally, the stopping criteria designed for this algorithm compares the non dominated solutions obtained so far at each generation with the solutions from the previous one. If the solutions from a previous generation remains unchangeable for a number of generations, then the algorithm will stop and return these solutions as an approximation of the \gls*{pof}.

\section{Experimental Results}\label{experimentalresults}

In these experiments, the main goal is to test which of the strategies proposed (and previously defined in Section 3.3) best fits for each weighted random functions (NUS, DUS and DGS) used in a particular set of mission planning problems. The experiments are divided in two parts: first, the different NUS, DUS and DGS strategies are applied and compared independently, i.e. the different NUS strategies are compared between them to get the best one, as well as the different DUS and the different DGS strategies. These strategies are compared against the previous approach, which is represented as the \textit{constant strategy}. Then, the best of these strategies for each part are selected and joined to obtain the best combination of strategies. This combination is tested and compared against the previous results.

For this purpose, a set of 16 missions of different complexity has been designed. These datasets are presented in Table \ref{tab:datasets}, where the different number of tasks (and Multi-UAV tasks), \gls*{uav2}, \gls*{gcs}, \gls*{nfz} and time dependencies are shown. 

The setup parameters of the \gls*{wrnsga2} are as follows. The selection criteria ($\mu + \lambda$) used was $30 + 300$, where $\lambda$ is the offspring size, and $\mu$ the elitism size, i.e. the number of the best parents that survive from the current generation to the next one. The mutation probability is $10\%$, and the number of generations used in the stopping criteria is 10. Each problem is run 10 times, and the average and standard deviation of the hypervolume~\cite{Zitzler2007} of the obtained solutions are calculated. The hypervolume has been computed using the normalized values of the objectives variables. Moreover, the number of generations needed to converge are extracted for each case. Instead of studying the computation time directly, which depends on the hardware and programming language used, the number of generations needed to converge, (which are proportional to the computation time) are studied in these experiments.

\begin{table}[!h]
\caption{Features of the different datasets designed.}
\label{tab:datasets}
\begin{center}
\scalebox{0.72}{
\begin{tabular}{|c!{\vrule width 3pt}c|c|c|c|c|c| }
  \hline
  \begin{minipage}[t]{1cm}
  \centering
  Dataset \\
  Id.
  \end{minipage} & Tasks & \begin{minipage}[t]{1.8cm}
  \centering
  Multi-UAV \\
  Tasks
  \end{minipage} & UAV & GCSs & NFZs & \begin{minipage}[t]{2cm}
  \centering
  Time \\
  Dependencies
  \end{minipage}\\[3ex]
  \noalign{\hrule height 2pt}	
  1 & 5 & 0 & 3 & 1 & 0 & 0 \\
  \hline
  2 & 6 & 1 & 3 & 1 & 1 & 0 \\
  \hline
  3 & 6 & 1 & 4 & 2 & 2 & 1 \\
  \hline
  4 & 7 & 1 & 5 & 2 & 1 & 2 \\
  \noalign{\hrule height 2pt}
  5 & 8 & 2 & 5 & 2 & 3 & 1 \\
  \hline
  6 & 9 & 2 & 5 & 2 & 0 & 2 \\
  \hline
  7 & 9 & 2 & 6 & 2 & 2 & 2 \\
  \hline
  8 & 10 & 2 & 6 & 2 & 3 & 3 \\
  \noalign{\hrule height 2pt}
  9 & 11 & 3 & 6 & 2 & 3 & 2 \\
  \hline
  10 & 12 & 3 & 7 & 3 & 0 & 2 \\
  \hline
  11 & 12 & 3 & 8 & 3 & 2 & 3 \\
  \hline
  12 & 13 & 4 & 7 & 3 & 4 & 4 \\
  \noalign{\hrule height 2pt}
  13 & 14 & 4 & 8 & 3 & 0 & 3 \\
  \hline
  14 & 15 & 4 & 9 & 3 & 5 & 4 \\
  \hline
  15 & 16 & 4 & 9 & 3 & 4 & 4 \\
  \hline
  16 & 16 & 5 & 10 & 3 & 5 & 5 \\
  \hline
\end{tabular}
}
\end{center}
\end{table}

For the sake of validating the statistical significance of the obtained results, the distributions of the metric values obtained by the different methods on each dataset have been compared by means of a nonparametric Kruskal-Wallis test \cite{Hollander2014}. This test represents the nonparametric version of the classical one-way ANOVA and is an extension of the Wilcoxon rank sum test to groups larger than 2. To this end, the test compares the medians of the group, and returns the $P$ value for the null hypothesis that all samples are drawn from the same population (or equivalently, from different populations with the same distribution). If the $P$ value is lower than a predefined $\alpha$, we can infer that the null hypothesis does not hold, that is, at least one sample median in the group is significantly different from the others, with $(1 - \alpha)$ level of confidence, then to determine which sample medians are statistically different,we have applied this multiple comparison procedure with $\alpha = 0.05$ (thus, with a $95\%$ level of confidence).

In the first experiment, a comparative assessment of the different strategies for the NUS weighted random function has been carried out, as shown in Table \ref{tab:resultsNUS}. These results show that the new strategies for the NUS weighted random function help reducing the number of generations needed to converge while maintaining the hypervolume of the solutions. Concretely, the effects of these strategies are specially appreciated in datasets with a bigger number of Multi-UAV tasks. The statistical significance of the results could be checked through the Kruskal-Wallis test, showing a significant relevance ($p<0.05$) for the results obtained in terms of number of generations needed to converge for datasets 4-9, while a significant relevance is provided for datasets 10-14 in terms of hypervolume. In datasets 1, the results are quite the same for the different strategies, so they are irrelevant, as datasets 2 and 3, which did not reach a significance relevance through the Kruskall-Wallis test. Datasets 15 and 16, on the other hand, did not reach any solution. It can be also appreciated as dataset 14, which did not find any solution using the common \gls*{nsga2} (constant strategy), could find solutions in some of the executions with the harmonic and geometric strategies. In general, the geometric strategy got better results in terms of number of generations needed to converge (it was the fastest approach), so we selected this strategy for the NUS weighted random function.
\begin{table}[!h]
\caption{Comparative assessment of the different weighted strategies used in NUS. Results show the average $\pm$ standard deviation for the hypervolume and the number of generations. For each dataset the best obtained result is marked in bold.}
\label{tab:resultsNUS}
\begin{center}
\scalebox{0.65}{
\begin{tabular}{ |c!{\vrule width 3pt}c|c!{\vrule width 2pt}c|c!{\vrule width 2pt}c|c!{\vrule width 2pt}c|c| }
  \hline
  Id. & \multicolumn{2}{c!{\vrule width 2pt}}{Constant Strategy} & \multicolumn{2}{c!{\vrule width 2pt}}{Arithmetic Strategy} & \multicolumn{2}{c!{\vrule width 2pt}}{Harmonic Strategy} & \multicolumn{2}{c|}{Geometric Strategy} \\
  \cline{2-9}
   & Hypervol. & No. Gen. & Hypervol. & No. Gen. & Hypervol. & No. Gen. & Hypervol. & No. Gen. \\
  \noalign{\hrule height 2pt}
  1 & $\mathbf{0.80 \pm 1e^{-6}}$ & $41 \pm 6$ & $0.80 \pm 1e^{-6}$ & $46 \pm 7$ & $0.80 \pm 1e^{-6}$ & $\mathbf{41 \pm 5}$ & $0.80 \pm 1e^{-6}$ & $44 \pm 7$ \\
  \hline
  2 & $0.72 \pm 1e^{-3}$ & $253 \pm 19$ & $0.72 \pm 2e^{-3}$ & $243 \pm 25$ & $\mathbf{0.72 \pm 1e^{-3}}$ & $\mathbf{231 \pm 23}$ & $0.72 \pm 3e^{-3}$ & $233 \pm 17$ \\
  \hline
  3 & $0.69 \pm 0.01$ & $395 \pm 26$ & $0.68 \pm 0.03$ & $382 \pm 27$ & $\mathbf{0.69 \pm 0.01}$ & $368 \pm 30$ & $0.69 \pm 0.02$ & $\mathbf{363 \pm 25}$ \\
  \hline
  4 & $0.74 \pm 0.02$ & $593 \pm 46$ & $0.74 \pm 0.03$ & $578 \pm 35$ & $0.73 \pm 0.04$ & $543 \pm 37$ & $\mathbf{0.74 \pm 0.01}$ & $\mathbf{537 \pm 21}$ \\
  \hline
  5 & $0.68 \pm 0.04$ & $408 \pm 32$ & $0.67 \pm 0.06$ & $393 \pm 36$ & $0.68 \pm 0.03$ & $376 \pm 29$ & $\mathbf{0.69 \pm 0.02}$ & $\mathbf{370 \pm 31}$ \\
  \hline
  6 & $0.63 \pm 0.04$ & $697 \pm 38$ & $0.63 \pm 0.04$ & $654 \pm 39$ & $0.63 \pm 0.05$ & $626 \pm 36$ & $\mathbf{0.63 \pm 0.03}$ & $\mathbf{601 \pm 38}$ \\
  \hline
  7 & $0.57 \pm 0.08$ & $865 \pm 41$ & $\mathbf{0.57 \pm 0.07}$ & $812 \pm 39$ & $0.57 \pm 0.08$ & $777 \pm 36$ & $0.57 \pm 0.07$ & $\mathbf{759 \pm 37}$ \\
  \hline
  8 & $0.40 \pm 0.10$ & $968 \pm 31$ & $0.42 \pm 0.08$ & $928 \pm 35$ & $0.43 \pm 0.10$ & $880 \pm 33$ & $\mathbf{0.43 \pm 0.08}$ & $\mathbf{846 \pm 26}$ \\
  \hline
  9 & $0.26 \pm 0.12$ & $1000 \pm 1$ & $0.31 \pm 0.14$ & $998 \pm 3$ & $0.33 \pm 0.13$ & $976 \pm 15$ & $\mathbf{0.33 \pm 0.11}$ & $\mathbf{951 \pm 18}$ \\
  \hline
  10 & $0.15 \pm 0.08$ & $1000 \pm 0$ & $0.23 \pm 0.14$ & $1000 \pm 1$ & $0.26 \pm 0.10$ & $997 \pm 2$ & $\mathbf{0.28 \pm 0.09}$ & $\mathbf{992 \pm 4}$ \\
  \hline
  11 & $0.05 \pm 0.06$ & $1000 \pm 0$ & $0.12 \pm 0.10$ & $1000 \pm 0$ & $0.19 \pm 0.09$ & $1000 \pm 0$ & $\mathbf{0.24 \pm 0.04}$ & $1000 \pm 0$ \\
  \hline
  12 & $1e^{-3} \pm 2e^{-3}$ & $1000 \pm 0$ & $0.03 \pm 0.01$ & $1000 \pm 0$ & $0.05 \pm 0.01$ & $1000 \pm 0$ & $\mathbf{0.11 \pm 0.06}$ & $1000 \pm 0$ \\
  \hline
  13 & $2e^{-5} \pm 3e^{-5}$ & $1000 \pm 0$ & $3e^{-3} \pm 3e^{-4}$ & $1000 \pm 0$ & $0.01 \pm 6e^{-3}$ & $1000 \pm 0$ & $\mathbf{0.06 \pm 0.01}$ & $1000 \pm 0$ \\
  \hline
  14 & $0 \pm 0$ & $1000 \pm 0$ & $0 \pm 0$ & $1000 \pm 0$ & $2e^{-3}  \pm 7e^{-4}$ & $1000 \pm 0$ & $\mathbf{0.01 \pm 2e^{-3}}$ & $1000 \pm 0$ \\
  \hline
  15 & $0 \pm 0$ & $1000 \pm 0$ & $0 \pm 0$ & $1000 \pm 0$ & $0 \pm 0$ & $1000 \pm 0$ & $0 \pm 0$ & $1000 \pm 0$ \\
  \hline
  16 & $0 \pm 0$ & $1000 \pm 0$ & $0 \pm 0$ & $1000 \pm 0$ & $0 \pm 0$ & $1000 \pm 0$ & $0 \pm 0$ & $1000 \pm 0$ \\
  \hline
\end{tabular}
}
\end{center}
\end{table}

We next proceed by analyzing the results of the experiments performed with the different strategies for the DUS weighted random function, which are presented in Table \ref{tab:resultsDUS}.
\begin{table}[!h]
\caption{Comparative assessment of the different weighted strategies used in DUS. Results show the average $\pm$ standard deviation for the hypervolume and the number of generations. For each dataset the best obtained result is marked in bold.}
\label{tab:resultsDUS}
\begin{center}
\scalebox{0.65}{
\begin{tabular}{ |c!{\vrule width 3pt}c|c!{\vrule width 2pt}c|c!{\vrule width 2pt}c|c!{\vrule width 2pt}c|c| }
  \hline
  Id. & \multicolumn{2}{c!{\vrule width 2pt}}{Constant Strategy} & \multicolumn{2}{c!{\vrule width 2pt}}{Arithmetic Strategy} & \multicolumn{2}{c!{\vrule width 2pt}}{Harmonic Strategy} & \multicolumn{2}{c|}{Geometric Strategy} \\
  \cline{2-9}
   & Hypervol. & No. Gen. & Hypervol. & No. Gen. & Hypervol. & No. Gen. & Hypervol. & No. Gen. \\
  \noalign{\hrule height 2pt}
  1 & $0.80 \pm 1e^{-6}$ & $41 \pm 6$ & $0.80 \pm 2e^{-6}$ & $40 \pm 6$ & $\mathbf{0.80 \pm 1e^{-6}}$ & $\mathbf{39 \pm 5}$ & $0.80 \pm 2e^{-6}$ & $39 \pm 6$ \\
  \hline
  2 & $0.72 \pm 1e^{-3}$ & $253 \pm 19$ & $0.72 \pm 2e^{-3}$ & $249 \pm 25$ & $\mathbf{0.72 \pm 1e^{-3}}$ & $240 \pm 23$ & $0.72 \pm 3e^{-3}$ & $\mathbf{238 \pm 17}$ \\
  \hline
  3 & $0.69 \pm 0.01$ & $395 \pm 26$ & $0.69 \pm 0.02$ & $366 \pm 24$ & $\mathbf{0.69 \pm 0.01}$ & $354 \pm 18$ & $0.68 \pm 0.03$ & $\mathbf{345 \pm 17}$ \\
  \hline
  4 & $0.74 \pm 0.02$ & $593 \pm 46$ & $0.73 \pm 0.04$ & $566 \pm 29$ & $\mathbf{0.74 \pm 0.02}$ & $531 \pm 33$ & $0.73 \pm 0.02$ & $\mathbf{519 \pm 25}$ \\
  \hline
  5 & $\mathbf{0.68 \pm 0.04}$ & $408 \pm 32$ & $0.68 \pm 0.05$ & $365 \pm 30$ & $0.67 \pm 0.03$ & $351 \pm 26$ & $0.64 \pm 0.02$ & $\mathbf{342 \pm 25}$ \\
  \hline
  6 & $0.63 \pm 0.04$ & $697 \pm 38$ & $0.62 \pm 0.03$ & $621 \pm 29$ & $\mathbf{0.63 \pm 0.04}$ & $572 \pm 41$ & $0.61 \pm 0.03$ & $\mathbf{539 \pm 63}$ \\
  \hline
  7 & $0.57 \pm 0.08$ & $865 \pm 41$ & $\mathbf{0.57 \pm 0.07}$ & $789 \pm 40$ & $0.56 \pm 0.08$ & $751 \pm 34$ & $0.51 \pm 0.11$ & $\mathbf{710 \pm 72}$ \\
  \hline
  8 & $0.40 \pm 0.10$ & $968 \pm 31$ & $0.43 \pm 0.08$ & $879 \pm 37$ & $\mathbf{0.44 \pm 0.10}$ & $848 \pm 32$ & $0.40 \pm 0.13$ & $\mathbf{825 \pm 69}$ \\
  \hline
  9 & $0.26 \pm 0.12$ & $1000 \pm 1$ & $0.40 \pm 0.10$ & $975 \pm 15$ & $\mathbf{0.42 \pm 0.09}$ & $929 \pm 45$ & $0.38 \pm 0.15$ & $\mathbf{903 \pm 89}$ \\
  \hline
  10 & $0.15 \pm 0.08$ & $1000 \pm 0$ & $0.29 \pm 0.09$ & $996 \pm 3$ & $\mathbf{0.33 \pm 0.10}$ & $972 \pm 12$ & $0.29 \pm 0.15$ & $\mathbf{921 \pm 78}$ \\
  \hline
  11 & $0.05 \pm 0.06$ & $1000 \pm 0$ & $0.19 \pm 0.10$ & $1000 \pm 0$ & $\mathbf{0.25 \pm 0.08}$ & $1000 \pm 0$ & $0.21 \pm 0.13$ & $\mathbf{956 \pm 98}$ \\
  \hline
  12 & $1e^{-3} \pm 2e^{-3}$ & $1000 \pm 0$ & $0.12 \pm 0.12$ & $1000 \pm 0$ & $\mathbf{0.19 \pm 0.07}$ & $986 \pm 69$ & $0.15 \pm 0.13$ & $\mathbf{936 \pm 123}$ \\
  \hline
  13 & $0 \pm 3e^{-5}$ & $1000 \pm 0$ & $0.04 \pm 0.01$ & $1000 \pm 0$ & $\mathbf{0.10 \pm 0.02}$ & $972 \pm 54$ & $0.06 \pm 0.01$ & $\mathbf{902 \pm 92}$ \\
  \hline
  14 & $0 \pm 0$ & $1000 \pm 0$ & $5e^{-3} \pm 4e^{-3}$ & $997 \pm 25$ & $0.02 \pm 0.01$ & $982 \pm 68$ & $\mathbf{0.03 \pm 4e^{-3}}$ & $\mathbf{897 \pm 142}$ \\
  \hline
  15 & $0 \pm 0$ & $1000 \pm 0$ & $0 \pm 0$ & $1000 \pm 0$ & $1e^{-3} \pm 4e^{-3}$ & $967 \pm 101$ & $\mathbf{0.01 \pm 2e^{-3}}$ & $\mathbf{902 \pm 168}$ \\
  \hline
  16 & $0 \pm 0$ & $1000 \pm 0$ & $0 \pm 0$ & $1000 \pm 0$ & $0 \pm 0$ & $1000 \pm 0$ & $0 \pm 0$ & $1000 \pm 0$ \\
  \hline
\end{tabular}
}
\end{center}
\end{table}

As can be seen, these results show that applying any of the strategies highly outperformed the convergence of the algorithm due to the guidance of the search. The Kruskal-Wallis test confirmed the significance relevance of the different strategies compared to the common \gls*{nsga2} (constant strategy) in either terms of number of generations and/or hypervolume, except for dataset 16, which did not converge in any case. In the first datasets it can be appreciated that the geometric strategy attains the best results, but as the complexity of the problem increases, the hypervolume for this strategy gets lower, probably due to local minima. For this purpose the harmonic strategy gets better results, so it will be regarded as the best-performing strategy for the final experiment.

Then, we consider the DGS weighted random function. The results of the experiments performed with the different strategies for this function are presented in Table \ref{tab:resultsDGS}.
\begin{table}[!h]
\caption{Comparative assessment of the different weighted strategies used in DGS. Results show the average $\pm$ standard deviation for the hypervolume and the number of generations. For each dataset the best obtained result is marked in bold.}
\label{tab:resultsDGS}
\begin{center}
\scalebox{0.65}{
\begin{tabular}{ |c!{\vrule width 3pt}c|c!{\vrule width 2pt}c|c!{\vrule width 2pt}c|c!{\vrule width 2pt}c|c| }
  \hline
  Id. & \multicolumn{2}{c!{\vrule width 2pt}}{Constant Strategy} & \multicolumn{2}{c!{\vrule width 2pt}}{Arithmetic Strategy} & \multicolumn{2}{c!{\vrule width 2pt}}{Harmonic Strategy} & \multicolumn{2}{c|}{Geometric Strategy} \\
  \cline{2-9}
   & Hypervol. & No. Gen. & Hypervol. & No. Gen. & Hypervol. & No. Gen. & Hypervol. & No. Gen. \\
  \noalign{\hrule height 2pt}
  1 & $0.80 \pm 1e^{-6}$ & $\mathbf{41 \pm 6}$ & $0.80 \pm 2e^{-6}$ & $44 \pm 5$ & $0.80 \pm 1e^{-6}$ & $42 \pm 6$ & $\mathbf{0.80 \pm 1e^{-6}}$ & $45 \pm 5$ \\
  \hline
  2 & $0.72 \pm 1e^{-3}$ & $253 \pm 19$ & $\mathbf{0.72 \pm 1e^{-3}}$ & $256 \pm 15$ & $0.72 \pm 3e^{-3}$ & $\mathbf{251 \pm 18}$ & $0.71 \pm 2e^{-3}$ & $255 \pm 20$ \\
  \hline
  3 & $\mathbf{0.69 \pm 0.01}$ & $395 \pm 26$ & $0.68 \pm 0.01$ & $390 \pm 20$ & $0.69 \pm 0.02$ & $388 \pm 17$ & $0.69 \pm 0.02$ & $\mathbf{383 \pm 18}$ \\
  \hline
  4 & $0.74 \pm 0.02$ & $593 \pm 46$ & $\mathbf{0.74 \pm 0.01}$ & $591 \pm 40$ & $0.74 \pm 0.02$ & $573 \pm 38$ & $0.74 \pm 0.01$ & $\mathbf{560 \pm 31}$ \\
  \hline
  5 & $0.68 \pm 0.04$ & $408 \pm 32$ & $\mathbf{0.68 \pm 0.03}$ & $398 \pm 34$ & $0.68 \pm 0.04$ & $385 \pm 36$ & $0.67 \pm 0.04$ & $\mathbf{379 \pm 27}$ \\
  \hline
  6 & $0.63 \pm 0.04$ & $697 \pm 38$ & $\mathbf{0.63 \pm 0.03}$ & $676 \pm 38$ & $0.63 \pm 0.04$ & $641 \pm 36$ & $0.62 \pm 0.02$ & $\mathbf{620 \pm 43}$ \\
  \hline
  7 & $0.57 \pm 0.08$ & $865 \pm 41$ & $0.57 \pm 0.06$ & $812 \pm 35$ & $\mathbf{0.57 \pm 0.06}$ & $795 \pm 40$ & $0.55 \pm 0.07$ & $\mathbf{778 \pm 55}$ \\
  \hline
  8 & $0.40 \pm 0.10$ & $968 \pm 31$ & $0.41 \pm 0.10$ & $938 \pm 36$ & $\mathbf{0.43 \pm 0.09}$ & $901 \pm 40$ & $0.42 \pm 0.10$ & $\mathbf{889 \pm 47}$ \\
  \hline
  9 & $0.26 \pm 0.12$ & $1000 \pm 1$ & $0.30 \pm 0.09$ & $983 \pm 9$ & $\mathbf{0.32 \pm 0.08}$ & $962 \pm 15$ & $0.30 \pm 0.10$ & $\mathbf{946 \pm 30}$ \\
  \hline
  10 & $0.15 \pm 0.08$ & $1000 \pm 0$ & $0.26 \pm 0.09$ & $998 \pm 2$ & $\mathbf{0.29 \pm 0.08}$ & $986 \pm 9$ & $0.26 \pm 0.11$ & $\mathbf{942 \pm 56}$ \\
  \hline
  11 & $0.05 \pm 0.06$ & $1000 \pm 0$ & $0.17 \pm 0.10$ & $1000 \pm 0$ & $\mathbf{0.21 \pm 0.09}$ & $1000 \pm 0$ & $0.19 \pm 0.11$ & $\mathbf{976 \pm 89}$ \\
  \hline
  12 & $1e^{-3} \pm 2e^{-3}$ & $1000 \pm 0$ & $0.09 \pm 0.06$ & $1000 \pm 0$ & $\mathbf{0.13 \pm 0.08}$ & $1000 \pm 0$ & $0.10 \pm 0.10$ & $\mathbf{967 \pm 102}$ \\
  \hline
  13 & $0 \pm 3e^{-5}$ & $1000 \pm 0$ & $0.02 \pm 8e^{-3}$ & $1000 \pm 0$ & $\mathbf{0.07 \pm 0.01}$ & $1000 \pm 0$ & $0.03 \pm 5e^{-3}$ & $\mathbf{938 \pm 103}$ \\
  \hline
  14 & $0 \pm 0$ & $1000 \pm 0$ & $0 \pm 0$ & $1000 \pm 0$ & $0 \pm 0$ & $1000 \pm 0$ & $0 \pm 0$ & $1000 \pm 0$ \\
  \hline
  15 & $0 \pm 0$ & $1000 \pm 0$ & $0 \pm 0$ & $1000 \pm 0$ & $0 \pm 0$ & $1000 \pm 0$ & $0 \pm 0$ & $1000 \pm 0$ \\
  \hline
  16 & $0 \pm 0$ & $1000 \pm 0$ & $0 \pm 0$ & $1000 \pm 0$ & $0 \pm 0$ & $1000 \pm 0$ & $0 \pm 0$ & $1000 \pm 0$ \\
  \hline
\end{tabular}
}
\end{center}
\end{table}

In this case, the behaviour is similar to DUS strategies, where the geometric strategy is better for the first datasets, but for more complex datasets, it gets stuck in local minima. The Kruskal-Wallis test is once more applied to confirm the significance relevance of the different strategies compared to the common \gls*{nsga2} in either terms of number of generations and/or hypervolume, showing relevance in datasets 5-13. So, as before, the harmonic strategy gets better results for more complex datasets, so it will be taken as the better strategy for the final experiment.

In the final experiment, we took the best strategies obtained in the previous experiments (geometric for NUS, harmonic for DUS and harmonic for DGS) and used the all together in the algorithm. The results obtained are shown in Table \ref{tab:resultsComb}, where the average, standard deviation and p-value obtained in the Kruskall-Wallis test (in comparison with the common \gls*{nsga2} approach with no strategies applied) are presented for both the hypervolume obtained and the number of generations needed to converge.
\begin{table}[!h]
\caption{Results (average $/$ standard deviation $/$ Kruskall-Wallis p-value) obtained with the best combination of strategies (geometric NUS, harmonic DUS and harmonic DGS).}
\label{tab:resultsComb}
\begin{center}
\scalebox{0.8}{
\begin{tabular}{ |c||c|c|c|c|c|c|c|c| }
  \hline
  Id. & Hypervolume & No. Generations \\
  \noalign{\hrule height 2pt}
  1 & $0.80 / 1e^{-6} / 0.76$ & $39 / 7 / 0.31$ \\
  \hline
  2 & $0.71 / 4e^{-3} / 0.56$ & $203 / 12 / 0.06$ \\
  \hline
  3 & $0.69 / 0.03 / 0.65$ & $274 / 21 / 0.03$ \\
  \hline
  4 & $0.74 / 0.02 / 0.54$ & $486 / 24 / 0.04$ \\
  \hline
  5 & $0.68 / 0.05 / 0.48$ & $334 / 29 / 0.03$ \\
  \hline
  6 & $0.63 / 0.02 / 0.45$ & $519 / 37 / 0.03$ \\
  \hline
  7 & $0.56 / 0.10 / 0.54$ & $703 / 41 / 0.04$ \\
  \hline
  8 & $0.48 / 0.09 / 0.41$ & $808 / 34 / 0.04$ \\
  \hline
  9 & $0.51 / 0.08 / 0.36$ & $876 / 31 / 0.03$ \\
  \hline
  10 & $0.36 / 0.09 / 0.09$ & $916 / 26 / 0.04$ \\
  \hline
  11 & $0.28 / 0.12 / 0.05$ & $987 / 9 / 0.02$ \\
  \hline
  12 & $0.31 / 0.07 / 0.04$ & $998 / 3 / 4e^{-3}$ \\
  \hline
  13 & $0.26 / 0.03 / 0.04$ & $1000 / 1 / 7e^{-5}$ \\
  \hline
  14 & $0.29 / 0.01 / 0.04$ & $1000 / 0 / 1.0$ \\
  \hline
  15 & $0.16 / 5e^{-3} / 0.04$ & $1000 / 0 / 1.0$ \\
  \hline
  16 & $0.06 / 5e^{-3} / 0.04$ & $1000 / 0 / 1.0$ \\
  \hline
\end{tabular}
}
\end{center}
\end{table}

These results show a notable improvement in the reduction of the number of generations needed to converge for complex datasets, which reflect an increased convergence rate. In addition, some valid solutions were obtained for dataset 16, where other approaches did not reach any solution. As the stopping criteria used for convergence checks that the non-dominated solutions obtained remain unchangeable for 10 generations, it is common that for the most complex datasets, with a lot of solutions in the \gls*{pof}, the algorithm did not reach the complete \gls*{pof} before the 1000 generations limit.

As there are 7 objectives to optimize, in order to visualize the form of the Pareto fronts obtained for each dataset, a parallel coordinates visualization \cite{Li2017} has been used to represent the solutions obtained. In order to ease the visualization and interpretation of the interplay between the different objectives, z-scores are used in these parallel plots, so changes are smoother depending on the standard deviation of the values for each objective.

\begin{figure}[!h]
        \includegraphics[scale=0.63]{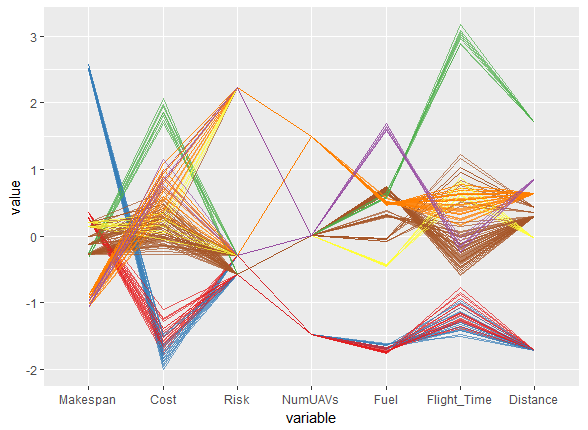}
		\centering
		\caption{Parallel Coordinates Visualization of the solutions obtained for dataset 4.}
		\label{fig:d4Parallel}
	\end{figure}

For instance, in Figure \ref{fig:d4Parallel}, the parallel coordinates visualization for dataset 4 is presented, where we can clearly distinguish several sets of solutions that have been clustered in different colors. In blue, it is appreciable that there are some solutions that have a high makespan, but instead have low values for the rest of variables. The red solutions are similar, but decreasing the makespan value while increasing the others except for the risk, fuel and number of \gls*{uav2}, which remain the lowest. Green solutions have the highest values for cost, flight time and distance, while preserving a medium value for makespan, fuel, risk and number of \gls*{uav2}. Purple solutions present the highest values for fuel and risk, but have the lowest values for makespan. Orange solutions also have low values for makespan and high values for risk and number of \gls*{uav2}, while the rest of variables have medium-high values. Finally, brown and yellow solutions have medium values for most variables, except for risk in some cases.
	\begin{figure}[!h]
        \includegraphics[scale=0.63]{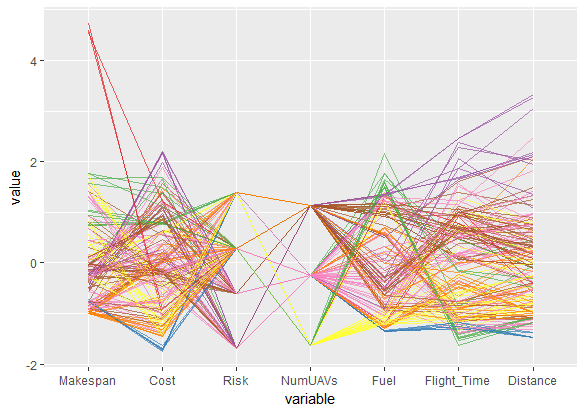}
		\centering
		\caption{Parallel coordinates visualization of the solutions obtained for dataset 9.}
		\label{fig:d9Parallel}
	\end{figure}
    
As the complexity of the problem increases, there are more non-dominated solutions and most times they are more scattered, so the visualization becomes harder. For example, Figure \ref{fig:d9Parallel} shows the parallel coordinates visualization for dataset 9. Here, it is more difficult to appreciate sets of solutions. In blue, some solutions with low cost, fuel and distance, but high risk and medium number of \gls*{uav2} are represented. Green solutions present a high value for fuel, while maintaining low values for flight time and number of \gls*{uav2}. Yellow solutions present a low number of \gls*{uav2} but a high risk. Purple solutions have high cost, flight time and distance, but low risk. Red solutions have the highest makespan, while orange solutions have the lowest, but the rest of variables are very scattered for these solutions.
	
\begin{figure}[!h]
		\includegraphics[scale=0.65]{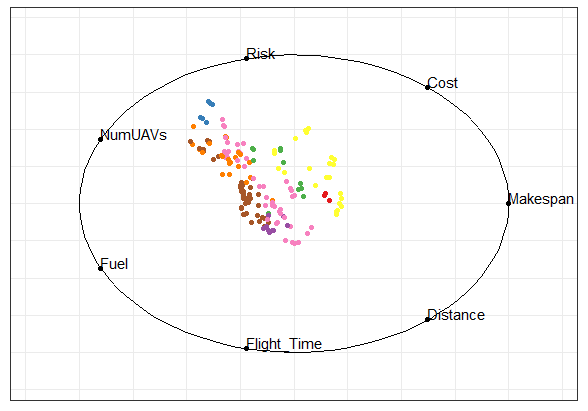}
		\centering
		\caption{Radial visualization (RadViz) of the solutions obtained for dataset 9.}
		\label{fig:d9Rad}
	\end{figure}

In these cases, sometimes is useful to use another type of visualization, such as radial visualization (RadViz) \cite{Tusar2015}, in order to visualize the clusterization of the solutions. The RadViz plot for dataset 9 can be seen in Figure \ref{fig:d9Rad}, maintaining the same colors used in Figure \ref{fig:d9Parallel}. In this plot, we can see that the same solutions clustered in the parallel coordinates approach are grouped together in RadViz (except for pink solutions, which represent the remaining solutions that were not clustered). In this visualization, we can appreciate how some sets of solutions approach to some objectives (when they have higher values) while they keep away from others (when they have lower values), e.g. blue solutions are very close to risk and number of \gls*{uav2}, because these solutions are worse for these variables, while they stay far from the rest of variables, were they performed better. In this sense, the solutions more centered in the graph, represent the ones having middle values for most variables, e.g. some pink solutions and the purple solutions. On the other hand, the yellow solutions keep away from the number of \gls*{uav2}, which are better optimized, but we can appreciate how some of them stay close to the cost, while other approach to the makespan, because they get different results in these two variables. Besides, the red solutions clearly approach the makespan, were they have the biggest value for this variable. It can be noticed that some color/solution pair sets are merged together, such as yellow and green solutions, or orange and brown solutions. This is due to the big similarities between these groups, were only one or two variable behave differently, and usually these variables still remain close, with no big numeric difference between them, e.g. green and yellow are mostly similar except for some yellow solutions that get better results in terms of fuel and cost, while green solutions do not.

With the combination of these two visualization methods, we can conclude that for complex datasets, as dataset 9, where there is a big number of solutions and they are highly sparsed, it is possible to appreciate some behaviors in terms of clustering of solutions, although most of these clusters are merged between them.

\section{Conclusions}\label{conclusions}

In this paper we have presented a new weighted random MOEA-CSP approach for solving the Multi-UAV mission planning problem. The presented approach considers missions consisting of several tasks to be performed by several \gls*{uav2} using a specific sensor. Each \gls*{uav2} is controlled by a \gls*{gcs} and use specific flight profiles. The problem has been modelled as a \gls*{csp} where the encoding of the \gls*{moea} considers the same variables as the \gls*{csp}. Thereby, the fitness function can use the \gls*{csp} to check if solutions are valid, and then a multi-objective fitness function optimizing several objectives is considered.


In addition, a biased random generator has been designed in order to guide the generation of new individuals. This generator improves the selection of the number of UAV for Multi-UAV tasks, and the selection of the UAV based on the distance to the task. For each selection, three strategies (arithmetic, harmonic and geometric) have been proposed to be used as weighted random functions.

The experiments performed over several missions show that this new approach outperforms the results obtained previously in terms of increased convergence rate. In addition, when the scenario of the mission is not favourable for the proposed weighted random approach, we obtain similar results to the previous approach. Moreover, this approach has resulted very helpful for problems of big complexity where convergence is difficult.

The main lack in this approach is the huge number of solutions obtained when missions are complex, which makes it difficult to select the final solution to be executed by the operator. Future works will focus on developing a \gls*{dss} for this problem, in order to select a few solutions among those obtained by the algorithm according to some quality metrics and the \gls*{gcs} operator profile.

\section{Acknowledgements}
This work has been co-funded by: Savier Project (Airbus Defence \& Space, FUAM-076915), Spanish Ministry of Science and Education and Competitivity (MINECO) and European Regional Development Fund (FEDER) under projects EphemeCH (TIN2014-56494-C4-4-P), and DeepBio (TIN2017-85727-C4-3-P), Comunidad Aut\'onoma de Madrid under project CIBERDINE S2013/ICE-3095, and the Basque Government for its support through the EMAITEK program. The authors would like to acknowledge the support obtained from Airbus Defence \& Space, specially from Savier Open Innovation project members: Jos\'e Insenser, C\'esar Castro and Gemma Blasco.

\section*{References}

\bibliography{mybibfile}

\end{document}